\documentclass[11pt]{article}

\usepackage[preprint]{acl}

\usepackage{times}
\usepackage{latexsym}
\usepackage{hyperref}
\usepackage{url}
\usepackage[table, dvipsnames]{xcolor}

\usepackage[utf8]{inputenc} %
\usepackage[T1]{fontenc}    %
\usepackage{hyperref}       %
\usepackage{url}            %
\usepackage{booktabs}       %
\usepackage{amsfonts}       %
\usepackage{nicefrac}       %
\usepackage{microtype}      %
\usepackage{xcolor}
\usepackage{tcolorbox}
\usepackage[table]{xcolor}
\usepackage{array}
\usepackage{rotating}  %

\usepackage{booktabs}
\usepackage{multirow}
\usepackage{tabularx}
\usepackage{amsmath}
\usepackage{graphicx}
\usepackage{wrapfig}
\usepackage{caption}
\usepackage{subcaption}
\usepackage{color,soul}
\usepackage{scalerel,graphicx,xparse}
\usepackage{float}
\usepackage{wrapfig}

\usepackage{algorithm}
\usepackage{algpseudocode}

\usepackage[export]{adjustbox}
\usepackage{enumerate}

\definecolor{myblue}{RGB}{114,166,202}
\definecolor{myred}{RGB}{224,71,76}

\usepackage{booktabs,arydshln}
\tcbuselibrary{listingsutf8}
\usepackage[T1]{fontenc}

\usepackage[utf8]{inputenc}

\usepackage{microtype}

\usepackage{inconsolata}

\usepackage{graphicx}

\title{Can VLMs Predict Future States? \\ Bootstrapping World Models from Inverse Dynamics}

\author{$^1$Yifu Qiu, $^{3,4}$Yftah Ziser, \\
$^2$\textbf{Anna Korhonen,} $^1$\textbf{Shay B. Cohen}, $^{1,2,3}$\textbf{Edoardo M. Ponti} \\
$^1$Institute for Language, Cognition and Computation, University of Edinburgh \\
$^2$Language Technology Lab, University of Cambridge \\
$^3$NVIDIA \\
$^4$University of Groningen \\
\texttt{\{yifu.qiu,scohen,eponti\}@ed.ac.uk} \\
}

\begin{document}
\maketitle
\begin{abstract}

Can unified vision--language models (VLMs) perform \textit{forward dynamics prediction} (FDP), i.e., predicting the future state (in image form) given the previous observation and an action (in language form)? We find that VLMs struggle to generate physically plausible transitions between frames from instructions. 
Nevertheless, we identify a crucial asymmetry in multimodal grounding: fine-tuning a VLM to learn \textit{inverse dynamics prediction} (IDP)—effectively captioning the action between frames—is significantly easier than learning FDP.
In turn, IDP can be used to bootstrap FDP through two main strategies: 1) weakly supervised learning from synthetic data and 2) inference time verification. Firstly, IDP can annotate actions for unlabelled pairs of video frame observations to expand the training data scale for FDP. 
Secondly, IDP can assign rewards to multiple samples of FDP to score them, effectively guiding search at inference time. We evaluate the FDP resulting from both strategies through the task of \textit{action-centric image editing} on \textsc{Aurora-Bench} with two families of VLMs. Despite remaining general-purpose, our best model achieves a performance competitive with state-of-the-art image editing models, improving on them by a margin between $7\%$ and $13\%$ according to GPT4o-as-judge, and achieving the best average human evaluation across all subsets of \textsc{Aurora-Bench}.\footnote{The code and models developed in this paper are available at \url{https://github.com/yfqiu-nlp/vlm-world-model}.}

\end{abstract}
\begin{figure}[t]
\centering
\includegraphics[width=\linewidth]{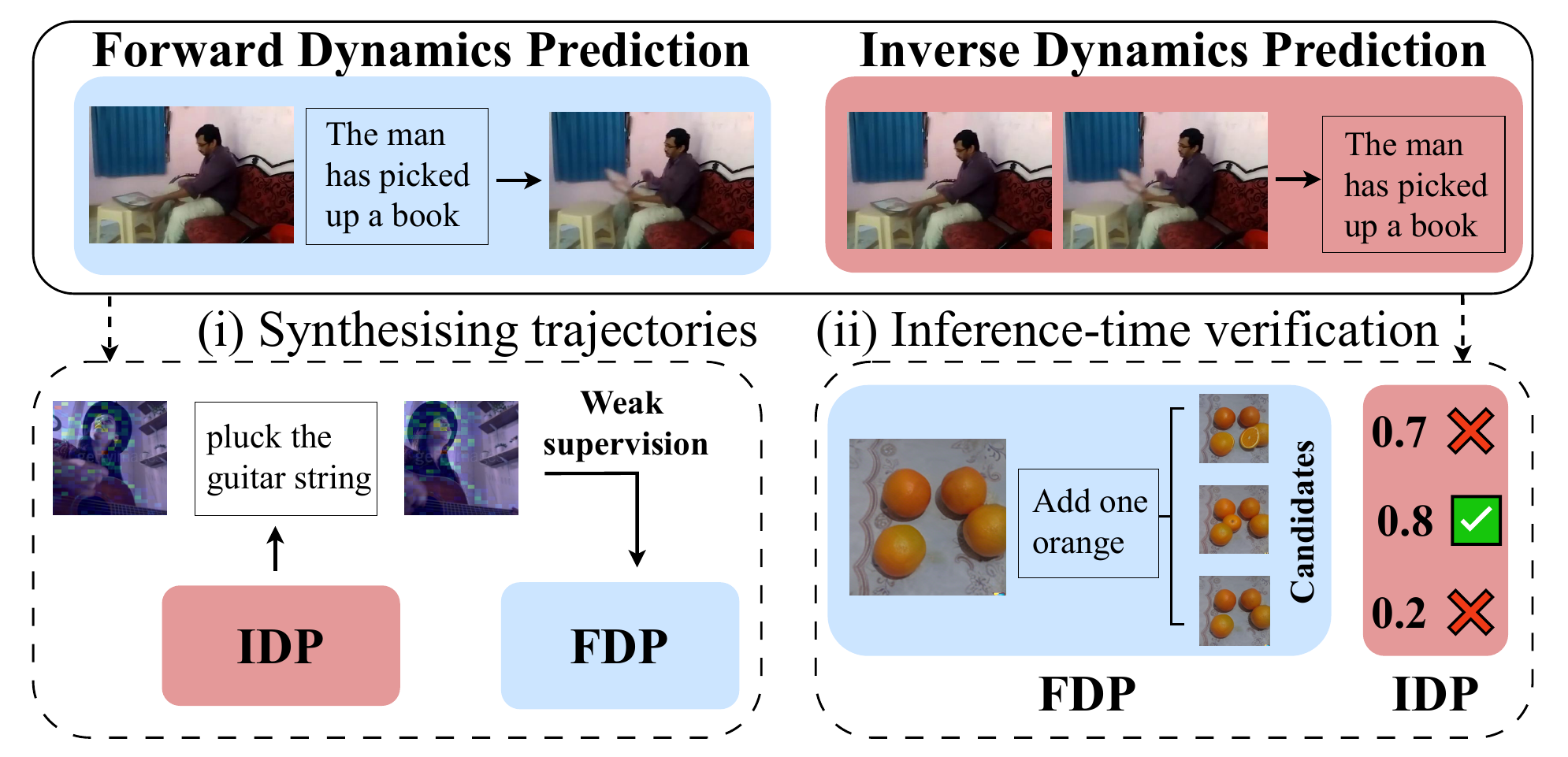}
\caption{A high-level illustration of our two strategies to bootstrap Forward Dynamics Prediction from Inverse Dynamics Prediction in unified Vision--Language Models: (i) synthesising trajectories for weak supervision (\textbf{left}) and (ii) inference-time verification of candidate future observations (\textbf{right}).}
\label{fig:main-figure-method}
\end{figure}
\section{Introduction}

\begin{figure*}[t]
    \centering
    \includegraphics[width=\linewidth]{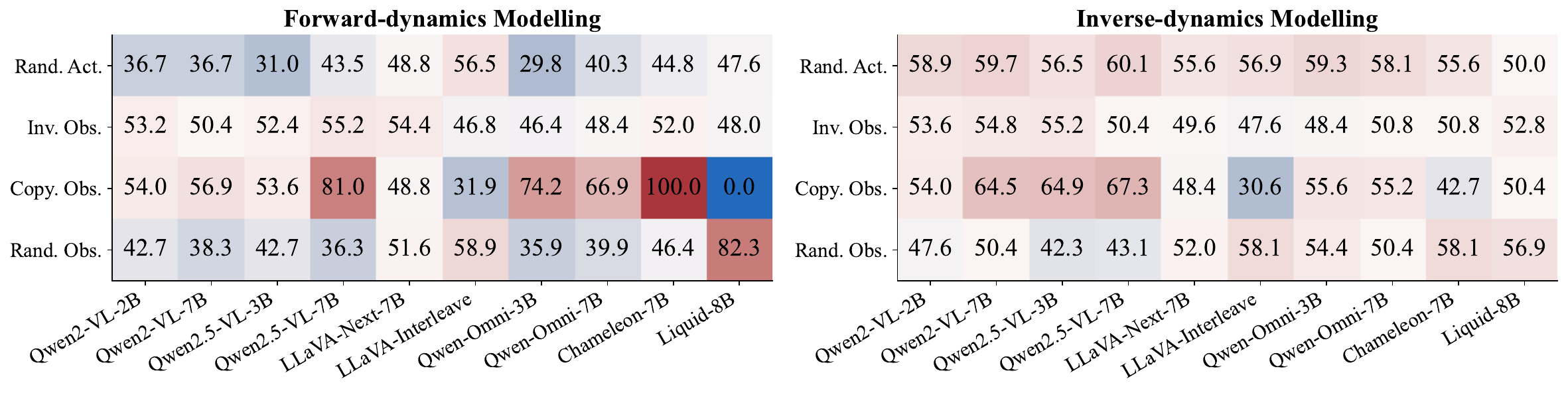}
    \caption{Percentage of times 9 VLMs assign higher probability to observation--action--observation Reference trajectories compared with 4 types of Negative (i.e., adversarially manipulated) trajectories, for both forward dynamics and inverse dynamics prediction. Higher values are better.}
    \label{fig:9-vlms-visualization}
\end{figure*}

\textit{World models} are instrumental in training embodied agents to endow them with specific abilities, such as planning and simulation \citep[\textit{inter alia}]{qin2024worldsimbench,brohan2023can,huanginner,li2025imagine,reed2022generalist, yang2023learning,Hafner2025}. However, learning a specialised world model is challenging due to the scarcity of real-world data \citep{liu2024world,motamed2025DoGenerativeModelsLearn}. %
Conversely, a promising alternative is endowing pre-existing unified\footnote{Here `unified' means models capable of interleaving text and images during generation architecturally.} vision-language models (VLMs) with world modelling abilities. %
In fact, VLMs are already imbued with plenty of real-world knowledge of both action (in language form) and perception (in vision form), because of their large-scale pre-training.

Firstly, we probe whether unified VLMs already contain reliable forward dynamics models (FDP), i.e., the ability to predict the next image frame given the previous image frame and an action expressed as a language instruction.
For this assessment, we limit ourselves to single-step trajectories, as a first step towards longer horizons. Based on our evaluation, we empirically demonstrate that existing VLMs do not prefer ground-truth trajectories compared to adversarially generated ones. Hence, we verify that the world model implicit in the original VLMs \textit{per se} is not well grounded on real-world forward dynamics \citep{gao2024physicallyGroundedVLM,qiu2024temporallyGrounded,abdou-etal-2021-language}.

Surprisingly, we also find that predicting the action $a \in \mathcal{A}$ taking place between observations $o \in \mathcal{O}$---also known as \textit{inverse dynamics prediction} (IDP;  $\mathcal{O}_t \times \mathcal{O}_{t+1} \rightarrow \mathcal{A}_t$)---via supervised fine-tuning is substantially easier than FDP ($\mathcal{O}_t \times \mathcal{A}_t \rightarrow \mathcal{O}_{t+1}$). Inspired by this finding, we propose two strategies to bootstrap the FDP from the IDP in a unified VLM, namely (i) \textbf{learning from synthetic trajectories} in videos automatically labelled with actions by the dynamics model; and (ii) \textbf{test-time verification} of predicted observations sampled from the FDP through the IDP. Figure~\ref{fig:main-figure-method} illustrates our two strategies.

For the weak supervision strategy, we use a VLM fine-tuned for IDP on the \textsc{Aurora} dataset \cite{krojer2024aurora} to annotate with linguistic actions ($\hat{a} \in \mathcal{A}$) motion key-frame pairs extracted from 45 hours of unlabelled real-world videos.
These are sourced from movements-in-time \citep{monfortmoments-MIT}, Kinetics700 \citep{kinetics700, carreira2019kinetics700Note} and UCF-101 \citep{soomro2012ucf101}.
Together with the ground-truth trajectories in \textsc{Aurora}, the synthesised trajectory triplets are then used for weakly supervised fine-tuning of the very same VLM for FDP. 
To encourage training to focus on image regions that change the most, we additionally propose a \textit{loss-weighting method} which weights the loss of each image token according to the visual difference between the ground-truth previous and next observations, as estimated by a recognition model.
Instead, in the verification strategy, we use the IDP to assign scores to multiple candidate samples generated by the FDP. Selecting the highest-score prediction can effectively guide search at inference time.

We conduct a thorough evaluation of both strategies to bootstrap FDP from IDP on several datasets from \textsc{Aurora-Bench} \citep{krojer2024aurora}: MagicBrush, Action-Genome, Something-Something, WhatsUp and Kubric. We experiment with two families of unified VLMs for FDP, Chameleon-7B and Liquid-8B. Training on trajectories synthesised by IDP, our FDP can achieve an overall performance superior to state-of-the-art diffusion models specialised for image editing, both in terms of GPT4o-as-a-judge and human evaluation. 
Inference-time verification can also improve FDP to a comparable degree as trajectory synthesis, providing an effective training-free bootstrapping method. %

While limited to single-step action--observation trajectories, this work offers promising early evidence that unified VLMs can be successfully endowed with FDP and hence may be suitably developed into long-horizon world models in the future.

\section{Unified VLMs Lack a Consistent Preference for Real-World Trajectories}
\label{sec:analysis-chameleon-zs-likelihood}
To understand whether off-the-shelf VLMs are already suitable for FDP, the first research question we investigate in this paper is: \textbf{To what extent do VLMs exhibit a preference for sequences of actions and observations that align with real-world trajectories?}

To address this question, we evaluate 9 different VLMs on ground-truth trajectories from 5 subsets of \textsc{Aurora-Bench} \citep{krojer2024aurora}: MagicBrush, Something-Something, Action-Genome, Whatsup, and Kubric.
Each subset contains 50 trajectory triplets of the form ($o_s, a, o_t$), where $o_s$ is the source observation (image frame), $a$ the action (text), and $o_t$ the target observation.\footnote{We choose these 9 VLMs using the following criteria: 1) they are publicly accessible and 2) they have been exposed to interleaved multimodal data during their pre-training.}

We then manually curate four types of negative trajectories using rule-based manipulations.
The first kind is an action-level manipulation. 
1) \textbf{Random Action}: for a given pair of observations, we substitute the original action with another randomly sampled within the same subset. 
We also perform three observation-level manipulations.
2) \textbf{Random Observation}: we randomly substitute the target observation with another in the same subset. 
3) \textbf{Copy Observation}: we copy the source observation as the target observation. 
4) \textbf{Inverse Observation}: we swap the source and target observations.

 In Figure~\ref{fig:9-vlms-visualization}, we compare the log-likelihood that VLMs assign to each ground-truth trajectory (Reference) against its corresponding manipulated one (Negatives). We evaluate the VLMs in two tasks: action prediction (i.e., as an inverse-dynamics model) and next-observation prediction (i.e., as a forward-dynamics model). For each kind of negative trajectory, we report the percentage of samples where the model favours the reference trajectory over the negative trajectory. From Figure~\ref{fig:9-vlms-visualization}, it emerges that VLMs display no clear preference for the ground-truth trajectories in a zero-shot setting (around $50\%$). 
 
 In the action prediction task (right panel), there is a slightly higher tendency to favour the ground-truth over the group with random actions; however, even in the best case, Qwen2.5-VL-7B prefers the reference in only 60.08\% of the samples. The only negative group that seems to be identifiable for VLMs is observation copying, where Qwen2.5-VL-7B has 67.34\% of correct preference.
In the next-observation prediction task (left panel), the VLM mostly fails in effectively differentiating the ground truth from the negatives. 
Although the underlying reason remains uncertain, one plausible explanation for this behaviour is that the model's ability to solve next-observation prediction tasks depends on their alignment with training sequences: for instance, it is plausible that Chameleon's data rarely features two identical adjacent images. 
We provide a discussion breaking down Chameleon's performance in Appendix~\ref{appendix:detailed-likelihood-anlaysis}.
 
\section{Bootstrapping FDP from IDP}
\label{sec:methodology}

Overall, the results from Section~\ref{sec:analysis-chameleon-zs-likelihood} reveal that off-the-shelf VLMs are not suitable for FDP and IDP \textit{per se}; however, they also show that IDP is a more feasible task, as VLMs already achieve accuracies above random chance in a zero-shot setting. Possibly, IDP abilities may be even improved with a small amount of fine-tuning. This underlies the key intuition behind our work: \textbf{can we leverage this asymmetry to bootstrap forward-dynamics abilities from inverse-dynamics ones within a unified VLM?}

We propose two strategies to leverage VLMs fine-tuned for IDP to enhance their own FDP abilities: (i) generating synthetic trajectories by annotating large-scale key-frame pairs from videos with actions predicted by IDP, then using these as weak supervision to train for FDP (Section~\ref{sec:synthetic-trajectories-creation}); and (ii) using the IDP as a verifier at test time to score candidate next observations sampled from the FDP (Section~\ref{sec:test-time-verification}). 

\begin{figure*}
    \centering
    \includegraphics[width=\linewidth]{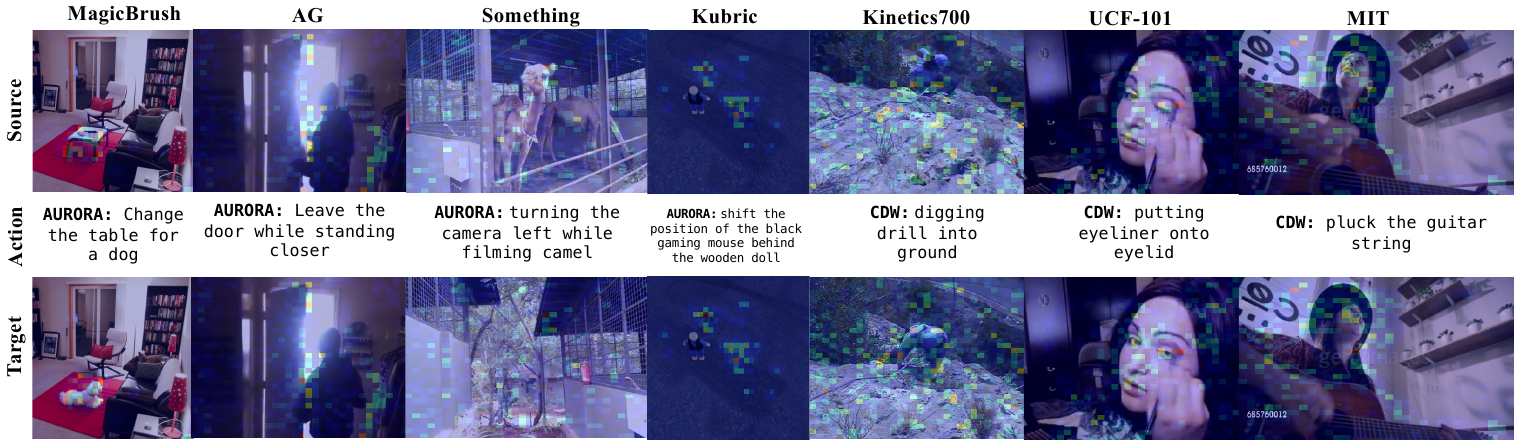}
    \caption{Heatmap visualization of image token weights predicted by the recognition model on examples from AG, Something-Something, MagicBrush, and Kubric, and UCF-101, Kinetics700 and MIT.}
    \label{fig:cog-loss-example}
\end{figure*}

\subsection{Inverse-dynamics Modelling}
\label{ssec:SFT}
First, we fine-tune the unified VLM as an inverse-dynamics model (\textbf{IDM}) $p_{\text{IDM}}(a\mid o_s, o_t)$, which predicts the probability of an action given the previous and next observations. As training data, we rely on high-quality triplets from \textsc{Aurora} \citep{krojer2024aurora} and the action recognition track of EPIC-Kitchen \citep{damen2018epic-kitchens}, which is based on videos with an egocentric view. We use the first and last frame in the EPIC-Kitchen video clips as the source and target observation $o_s$ and $o_t$ and the annotated action as $a$. We provide full details on IDM training data and setting in Appendix~\ref{appendix:training-details-IDM}.

\subsection{Weakly Supervised Learning from Unlabelled Videos} 
\label{sec:synthetic-trajectories-creation}
\paragraph{Synthetic Trajectories.} Taking advantage of the resulting high-quality IDM, we then explore the first of our strategies to bootstrap the FDP in VLMs:  
we annotate pairs of motion key-frames of unlabelled videos with a textual description of the action with the IDM. To ensure both scale and quality, we collect approximately 45 hours of video from Moments-in-Time \citep{monfortmoments-MIT}, Kinetics-700 \citep{kinetics700, carreira2019kinetics700Note}, and UCF-101 \citep{soomro2012ucf101}, all of which consist of curated clips focused on human actions. To ensure the selected pairs of motion key-frames are meaningful, i.e., they express a valid action, we then calculate the optical flow to quantify the dynamics per frame in the video clips, and select the top-$K_{f}$ frames while ensuring that the interval between two selected frames is $I_f$. Specifically, we set $I_f=20$ and $K_f=6$ for all three datasets. This results in approximately 20K, 46K, and 21K annotated trajectory triplets from Moments-in-Time, Kinetics-700, and UCF-101, respectively.
Finally, we apply a filtering strategy to further guarantee the quality of the resulting triplets. Specifically, we apply stratified top-$k$ sampling based on the IDM's predicted likelihood for each trajectory triplet ($o_s$, $\hat{a}_{\text{IDM}}$, $o_t$)\footnote{The details of this algorithm are provided in Appendix~\ref{appendix:stratified-top-k}.} to select a subset of triplets. 
We show statistics of the scores and action classes for the selected triplets in Figure~\ref{fig:scores-per-action-class}. We also provide one example for each dataset in Figure~\ref{fig:cog-loss-example}.

\paragraph{FDP.} Finally, we fine-tune a VLM as a forward-dynamics model (\textbf{FDM}), $p_\text{FDM}(o_t \mid a, o_s)$ on both \textsc{Aurora}’s supervised triplets $\mathcal{D}_{\text{sup}}$ and unsupervised triples $\mathcal{D}_{\text{unsup}}$ with actions sampled from the IDM. Normally, a FDM would be trained with maximum likelihood estimation as an objective:
\begin{equation}
\label{eq:simple_loss}
\begin{aligned}
\small
&\min_{\theta}\; 
\mathbb{E}_{(a, o_s, o_t) \sim \mathcal{D}_{\text{sup}}}
\left[ -\log p_{\theta}(o_t \mid a, o_s) \right] + \\
&\mathbb{E}_{(o_s, o_t) \sim \mathcal{D}_{\text{unsup}}}
\left[
\mathbb{E}_{\hat{a} \sim p_{\text{IDM}}(a \mid o_s, o_t)}
\left[ -\log p_{\theta}(o_t \mid \hat{a}, o_s) \right]
\right]
\end{aligned}
\end{equation}
where $\theta$ are the parameters for FDM, and $\hat{a}$ is an action sampled from the IDM.

\paragraph{Recognition-Weighted Training Loss.} 

Nevertheless, the objective in Equation~\ref{eq:simple_loss} is limited by treating all regions (i.e., image patches) of the target observation equally, even if some of them remain identical to the source whereas others change. This may result in degenerate solutions such as always copying the source.
As an alternative, we therefore propose a novel training objective for FDM that overcomes this assumption. This objective weights the loss of next-observation image tokens based on their importance. The intuition is that not all image patches in source and target observations contribute equally to modelling real-world transitions; instead, the model should focus on patches most indicative of the action's consequences. To this end, we leverage a recognition model $f_{\text{rec}}(w; o_s, o_t)$, which outputs token-level weights that represent the similarity between source and target patches. These weights modulate the loss during training, emphasising learning on semantically meaningful regions and down-weighting irrelevant ones. We formulate our objective as:
\begin{equation}
\small
\min_{\theta} \; \sum_{l=1}^{L} f_{\text{rec}}(w ; o_s, o_t)^{(l)} \cdot \left( -\log p_{\theta}(o_t^{(l)} \mid o_t^{(<l)}, o_s, a) \right),
\end{equation}
where $\theta$ are the parameters of VLM as FDM and a $L$ is the number of image tokens. $o_t^{(l)}$ and $o_t^{(<l)}$ represent the image tokens of $o_t$ at position $l$ and the history of previous positions, respectively. For simplicity, we use the pre-trained vector-quantised encoder of the unified VLMs as the recognition model, by computing the squared L$_2$ norm of pre-quantized features $\mathbf{z}_{o_s} \in Z_{o_s}$ and $\mathbf{z}_{o_t} \in Z_{o_t}$  where $Z_{o_s}$ and $Z_{o_t}$ are the sets of features of source and target observations, respectively. We visualise the token weights in Figure~\ref{fig:cog-loss-example}, which capture where acting on the source observation yields the largest effects on the target one.

\subsection{Test-time Verification}
\label{sec:test-time-verification}

Finally, we introduce an inference-time strategy which harnesses the IDM as a verifier to enhance FDM performance. Inspired by recent work on scaling test-time compute \citep{muennighoff2025s1,snell2024scaling}, we let the FDM generate $N$ candidate observations. Each candidate is paired with the source and scored by the IDM, which assigns each a predicted likelihood, interpreted as a reward. The final prediction of the FDM is selected by maximising the IDM's reward:
\[
\begin{aligned}
\hat{o}_t
&= \mathop{\mathrm{argmax}}_{i \in \{1, \dots, N\}}
p_{\text{IDM}}\!\left(a \mid o_s, o_t^{(i)}\right), \\
&\quad \text{where } 
o_t^{(i)} \sim p_{\text{FDM}}(o_t \mid o_s, a).
\end{aligned}
\]
where 
$\hat{o}_t$ is the selected prediction. While this strategy is training-free, it requires sampling multiple candidates at inference time.

\section{Experiments and Results}
\label{sec:experiments-results}

In Section~\ref{ssec:experiment_setting}, we first describe the experimental setup, including benchmarks, baselines, and evaluation metrics.
We then report results on the inverse dynamics prediction task in Section~\ref{ssec:idm_result} to verify that fine-tuning on a limited amount of examples is sufficient to develop robust IDP in a VLM.
This is followed in Section~\ref{ssec:fdm_result} by automatic and human evaluation of both strategies introduced in Section~\ref{sec:methodology} to bootstrap FDP, as well as ablation studies.
Finally, we analyse inference-time verification and examine the transfer of forward dynamics prediction to two spatial reasoning benchmark.

\subsection{Experimental Setting}
\label{ssec:experiment_setting}

\paragraph{Benchmarks.} We select \textsc{Aurora-Bench} \citep{krojer2024aurora} for evaluation of both IDM and FDM. This dataset provides high-quality data for action-centric edits, covering a wide array of phenomena and assessing a model's alignment with the physical world. We choose 5 subsets: \textbf{MagicBrush} for specialised image editing, \textbf{Action-Genome (AG)} and \textbf{Something-Something (Something)} for real-world actions. \textbf{Whatsup} focuses on spatial reasoning, whereas \textbf{Kubric} contains samples from a physical engine \citep{greff2022kubric}.

\paragraph{Models and Baselines.} We experiment with two unified VLMs, Chameleon-7B \citep{team2024chameleon} and Liquid-8B \citep{wu2024liquid}. The variants of these models fine-tuned on both supervised and weakly supervised data with loss weighting (i.e., with the training-time bootstrapping strategy) are indicated as \textbf{C-FDM} and \textbf{L-FDM}, respectively.
As our first baselines, we use the same two VLMs either zero-shot (\textbf{C-ZS} and \textbf{L-ZS}) or fine-tuned \textit{only} on \textsc{Aurora}'s supervised data (\textbf{C-FT} and \textbf{L-FT}). %
Additionally, we include three state-of-the-art diffusion models specialised for image editing as baselines, such as \textbf{PixInstruct} \citep{brooks2023instructpix2pix}, \textbf{GoT} \citep{fang2025got} and \textbf{SmartEdit} \citep{huang2024smartedit}. Finally, as a sanity check, we also report the metric scores obtained by simply copying the source observation input as the next-observation prediction (\textbf{Copy}).

\paragraph{Metrics.} For next-observation prediction (FDP) evaluation, following \citet{fang2025got}, we rely on \textbf{GPT4o-as-a-judge} as it is the only metric that reliably penalises Copy. In Appendix~\ref{appendix:all-5-metrics-results}, we show four other metrics, e.g., CLIP, which assign high scores to Copy. 
GPT4o-as-a-judge scores consider two criteria, one for the editing success rate and one for visual consistency with the original. We take the minimum of the two as the final score. The prompt for the judge is provided in Appendix~\ref{appendix:gpt4o-judge}.

\subsection{Inverse Dynamics Prediction}
\label{ssec:idm_result}

\begin{table}[t]

\centering
\footnotesize
\begin{tabular}{llllll}
\toprule
                            & \textbf{BS} & \textbf{R-1}  & \textbf{R-L} & \textbf{BLEU}  \\ \midrule
Chameleon ZS        & 0.05          & 0.09        & 0.08    & 0.00 \\
Chameleon IDM & \textbf{0.45}          & \textbf{0.45}    & \textbf{0.44}    & 0.20 \\
Liquid IDM & 0.41          & 0.41    & 0.36    & \textbf{0.21} 
\\ 
\bottomrule
\end{tabular}
\caption{Performance of Inverse Dynamics Models on action prediction, measured by text similarity metrics: BERTScore (BS; \citealt{zhangbertscore}), ROUGE (R-1, R-L; \citealt{lin2004rouge}) and BLEU \citep{papineni2002bleu}.}

\label{tab:action-prediction-text-similarity}
\end{table}

We evaluate the IDM based on the textual similarity of the predicted action with the ground-truth action in \textsc{Aurora-Bench}. Results are shown in Table~\ref{tab:action-prediction-text-similarity}. Our results demonstrate that a moderate amount of fine-tuning is necessary to let unified VLMs verbalise the dynamics connecting two observations, as evidenced by the large performance gap between the zero-shot and fine-tuned models.
It is also worth noting that both models' IDP performance (in terms of BLEU) is similar. We leverage these IDP versions of Chameleon and Liquid to bootstrap FDP, whose results are reported in Section~\ref{ssec:fdm_result}.

\subsection{Forward Dynamics Prediction}
\label{ssec:fdm_result}
Next, we report FDP results for each of the two bootstrapping strategies: in Section~\ref{ssec:res_syntra} for trajectory synthesis and in Section~\ref{sec:res_infver} for inference-time verification.

\subsubsection{Synthesising Trajectories with IDP}
\label{ssec:res_syntra}

\begin{table*}[t]
\resizebox{\textwidth}{!}{
\begin{tabular}{lccc|ccc|ccc}
\toprule
\multirow{2}{*}{\textbf{Dataset}} 
& \multicolumn{9}{c}{\textbf{Models}} \\
\cmidrule(lr){2-10}
& \textbf{PixInst} 
& \textbf{GoT} 
& \textbf{SE} 
& \textbf{C-ZS} 
& \textbf{C-FT} 
& \textbf{C-FDM} 
& \textbf{L-ZS} 
& \textbf{L-FT} 
& \textbf{L-FDM} \\
\midrule

\multicolumn{1}{l}{\textit{Forward-dynamics Prediction (GPT4o score)}} 
& \multicolumn{3}{c|}{} 
& \multicolumn{3}{c|}{} 
& \multicolumn{3}{c}{} \\

\textbf{MagicBrush}
& 3.12 & 5.96 & \textbf{6.71} & 0.00 
& \cellcolor{red!20}2.52 
& \cellcolor{blue!20}3.92
& 0.68 & \cellcolor{red!20}5.56 & \cellcolor{blue!20}5.82 \\

\textbf{AG}
& 1.20 & 1.61 & 3.08 & 0.17
& \cellcolor{red!20}2.48
& \cellcolor{blue!20}\textbf{3.64}
& 0.08 & \cellcolor{red!20}2.59 & \cellcolor{blue!20}2.98 \\

\textbf{Something}
& 0.96 & 2.62 & 2.81 & 0.37
& \cellcolor{blue!20}\textbf{3.11}
& \cellcolor{red!20}2.92
& 0.42 & \cellcolor{red!20}2.72 & \cellcolor{blue!20}2.88 \\

\textbf{WhatsUp}
& 0.00 & 1.58 & 0.76 & 0.15
& \cellcolor{blue!20}0.88
& \cellcolor{red!20}0.54
& 0.82 & \cellcolor{red!20}2.88 & \cellcolor{blue!20}\textbf{3.30} \\

\textbf{Kubric}
& 1.88 & 3.92 & 3.70 & 0.14
& \cellcolor{red!20}7.30
& \cellcolor{blue!20}\textbf{7.32}
& 2.22 & \cellcolor{red!20}6.28 & \cellcolor{blue!20}6.60 \\

\midrule
\textbf{\textsc{Aurora-Bench} Avg.}
& 1.43 & 3.14 & 3.41 & 0.17
& \cellcolor{red!20}3.26
& \cellcolor{blue!20}3.67
& 0.84 & \cellcolor{red!20}4.04 & \cellcolor{blue!20}\textbf{4.32} \\

\midrule

\multicolumn{1}{l}{\textit{Spatial Reasoning (Accuracy)}} 
& \multicolumn{3}{c|}{} 
& \multicolumn{3}{c|}{} 
& \multicolumn{3}{c}{} \\

\textbf{SpatialMQA}
& -- & -- & -- & 26.1
& 25.8 & {27.2}
& 27.7 & \textbf{28.0} & 27.8 \\

\textbf{EmbodiedSpatial-Bench}
& -- & -- & -- & 15.1
& {21.2} & 17.5
& 32.6 & 33.2 & \textbf{33.8} \\

\bottomrule
\end{tabular}
}
\caption{
Performance on \textsc{Aurora-Bench} (GPT-4o-as-a-judge score) and spatial reasoning benchmarks (accuracy).
We \textbf{bold} the best model per dataset.
Among our variants, we highlight the \colorbox{blue!20}{best} and \colorbox{red!20}{worst} scores.
SE: SmartEdit.
Greedy decoding is used for spatial reasoning evaluation and applied only to VLMs.
}
\label{tab:main-result}

\end{table*}

\paragraph{Automatic evaluation.} 
To test trajectory synthesis, we evaluate FDM on next-observation prediction for each of the \textsc{Aurora-Bench} subsets, reporting GPT4o-\-as-\-a-\-judge scores in Table~\ref{tab:main-result}.
We first notice that the state-of-the-art image editing models (i.e., PixInstruct, GoT, SmartEdit) tend to specialise on the proper image editing subset MagicBrush ($5.96$ and $6.71$ GPT4o scores for GoT and SmartEdit). Nevertheless, in the action-centric subsets, including Action-Genome (AG), Something and Kubric, they are mostly behind bootstrapped models (C/L-FDM) and even the fine-tuned baselines (C/L-FT). %
Crucially, considering the average performance of C/L-FDM on all subsets of \textsc{Aurora-Bench} reveals the benefit of augmenting the training data with synthetic triplets bootstrapped from the IDM (vs.\ FT), as it yields a 13\%  gain for Chameleon and 7\% for Liquid, and the benefit of fine-tuning off-the-shelf VLMs for FDP more broadly (vs.\ ZS). 

\begin{table}[t]
    \small
    \centering
        \begin{tabular}{l|cc|cc}
        \toprule
        \multicolumn{1}{c}{}                                   & \multicolumn{2}{c}{\textbf{Weighted}} & \multicolumn{2}{c}{\textbf{Standard}} \\ 
        \multicolumn{1}{c|}{}                                   & ES ($\uparrow$)              & ME ($\uparrow$)            & ES ($\uparrow$)                          & ME ($\uparrow$)                   \\ \midrule
        \textbf{MB}                                            & \cellcolor{blue!20}{3.73}            & \cellcolor{red!20}{8.17}           & \cellcolor{red!20}{3.68}                        & \cellcolor{blue!20}{8.46}                  \\
        \textbf{AG}                                            & \cellcolor{blue!20}{3.18}            & \cellcolor{red!20}{8.03}           & \cellcolor{red!20}{2.37}                        & \cellcolor{blue!20}{8.13}                  \\
        \textbf{ST}                                            & \cellcolor{blue!20}{3.32}            & \cellcolor{red!20}{7.01}           & \cellcolor{red!20}{2.78}                        & \cellcolor{blue!20}{7.20}                   \\
        \textbf{WU}                                            & \cellcolor{red!20}{0.54}            & \cellcolor{blue!20}{7.25}           & \cellcolor{blue!20}{0.76}                        & \cellcolor{red!20}{7.19}                  \\
        \textbf{KU}                                            & \cellcolor{blue!20}{7.75}            & \cellcolor{red!20}{8.49}           & \cellcolor{red!20}{7.24}                        & \cellcolor{blue!20}{8.70}                   \\ \midrule
        \textbf{Avg.}                                           & \cellcolor{blue!20}{3.71}            & \cellcolor{red!20}{7.80}            & \cellcolor{red!20}{3.37}                        & \cellcolor{blue!20}{7.94}                 \\ 
        \textbf{GPT4o} & \multicolumn{2}{c|}{\cellcolor{blue!20}{3.67}}  & \multicolumn{2}{c}{\cellcolor{red!20}{3.58}} \\
         \bottomrule
        \end{tabular}
        \caption{Detailed scores of GPT4o-as-a-judge evaluation for loss-weighting and standard training. We report the scores for \textbf{E}diting \textbf{S}uccess (\textbf{ES}) and \textbf{M}inimal \textbf{E}diting (\textbf{ME}). MB: MagicBrush, AG: Action-Genome, ST: Something-Something, WU: WhatsUp, KU: Kubric. We highlight the \colorbox{blue!20}{best} and \colorbox{red!20}{worst} scores for each category. We report the average of ES and ME as \textbf{Avg.} and the final score as \textbf{GPT4o}.}
        \label{tab:gpt4o-score-analysis}
\end{table}

\paragraph{Human Evaluation.} Following \citet{krojer2024aurora}, we conduct a blind human evaluation comparing GoT, SmartEdit, C-FT, and C-FDM. We randomly sample 5 examples from each subset within \textsc{Aurora-Bench} and present the outputs generated by each of the four models. Human annotators are asked to identify the best and worst generated observations based on three criteria: (1) \textit{Realism}: the generated image should exhibit natural textures and lighting while remaining faithful to the input scene; (2) \textit{Instruction-Following Ability}: the edit should clearly reflect the given action; and (3) \textit{Over-Editing}: the modification should be minimal and focused, altering only what is necessary to fulfil the action. Each model receives +1 point for being selected as the best, -1 for the worst, and 0 otherwise. We compute the average scores over 350 annotated samples, as reported in Table~\ref{tab:human-eval-results}. The results are well aligned with automatic evaluations in Table~\ref{tab:main-result}: image-editing models excel in the MagicBrush and WhatsUp subsets, but fall short on action-centric datasets such as Action-Genome, Something-Something, and Kubric. In contrast, C-FDM outperforms C-FT (and all other baselines) on all three of these datasets, highlighting its strength in next-observation prediction in real-world, action-centric trajectories despite remaining a general-purpose VLM.

\begin{table*}[t]
    \centering
    \begin{minipage}{0.48\textwidth}
        \centering
        
        \begin{tabular}{lccc}
        \toprule
                            & \textbf{C-FDM} & \textit{w/o Synth.} & \it{ w/o LW} \\ \midrule
        \textbf{MB} & \textbf{3.48}         & -0.28            & -0.22                 \\
        \textbf{AG}         & \textbf{3.02}         & -0.35            & -0.08                 \\
        \textbf{ST}  & \textbf{3.06}         & -0.18            & -0.19                 \\
        \textbf{WU}    & \textbf{0.46}         & \phantom{-}0.40             & \phantom{-}0.08                  \\
        \textbf{KU}  & \textbf{7.14}         & -0.03            & -0.33                 \\ \midrule
        \textbf{All}        & \textbf{3.43}         & -0.09            & -0.15                 \\ \bottomrule
        \end{tabular}
        \caption{Ablation study of synthetic trajectories (Synth.) and loss weighting (LW) in C-FDM. Numbers are GPT-4o-as-judge scores ($\uparrow$, average of 3 runs). MB: MagicBrush, AG: Action-Genome, ST: Something-Something, WU: WhatsUp, KU: Kubric.  }
        \label{tab:ablation-study}
    \end{minipage} \hfill
    \begin{minipage}{0.48\textwidth}
    \centering
        \begin{tabular}{lcccc}
        \toprule
                           & \textbf{GoT}  & \textbf{SE} & \textbf{C-FT} & \textbf{C-FDM}  \\ \midrule
        \textbf{MB}        & \phantom{-}0.06${^\dagger}$          & \phantom{-}\textbf{0.29}${^\dagger}$         & -0.32${^\dagger}$           & -0.03         \\
        \textbf{AG}        & -0.23${^\dagger}$         & -0.46${^\dagger}$                 & \phantom{-}0.32\phantom{$^\dagger$}            & \phantom{-}\textbf{0.37} \\
        \textbf{ST} & \phantom{-}0.00\phantom{$^\dagger$}          & -0.37${^\dagger}$                 & \phantom{-}0.18\phantom{$^\dagger$}           & \phantom{-}\textbf{0.20} \\
        \textbf{WU}   & \phantom{-}\textbf{0.25}\phantom{$^\dagger$} & -0.38${^\dagger}$                 & \phantom{-}0.14\phantom{$^\dagger$}            & \phantom{-}0.00          \\
        \textbf{KU}   & -0.52${^\dagger}$         & -0.22${^\dagger}$                 & \phantom{-}0.34\phantom{$^\dagger$}            & \phantom{-}\textbf{0.40} \\ \midrule
        \textbf{All}       & -0.09${^\dagger}$         & -0.23${^\dagger}$                 & \phantom{-}0.13\phantom{$^\dagger$}            & \phantom{-}\textbf{0.19} \\ \bottomrule
        \end{tabular}
        \caption{Human evaluation results. ${\dagger}$ indicates all results whose gap with respect to C-FDM is significant, based on a Wilcoxon signed-rank test ($p=0.05$). MB: MagicBrush, AG: Action-Genome, ST: Something-Something, WU: WhatsUp, KU: Kubric.}
        \label{tab:human-eval-results}
    \end{minipage}
\end{table*}

\paragraph{Reliability of GPT-4o-as-a-Judge.} A natural concern about our evaluation protocol is whether GPT-4o-as-a-judge faithfully reflects human preferences, and whether it may systematically favour its own family of VLM-generated outputs over specialised image-editing baselines. We first note that traditional similarity-based metrics commonly used in image editing, e.g., L1-distance, CLIP-I, CLIP-T, and DINO---systematically reward the degenerate \textit{Copy} baseline (see Appendix~\ref{appendix:all-5-metrics-results}, Table~\ref{tab:main-result-5-metrics}), which returns the source observation as the prediction without performing any edit. GPT-4o-as-a-judge, in contrast, assigns a score of $0$ to \textit{Copy} across every subset of \textsc{Aurora-Bench}, and is the same protocol adopted by recent state-of-the-art image editing works \citep{huang2024smartedit,fang2025got}. To further test whether GPT-4o tracks human preferences, we compute the Spearman rank correlation between human best/worst/rest annotations (encoded as $+1/-1/0$) and the corresponding GPT-4o scores across all four models (GoT, SmartEdit, C-FT, C-FDM). We find a statistically significant positive correlation ($\rho = 0.28$, $p < 0.001$); the value is deflated by the scale mismatch between the continuous GPT-4o scores and the trinary human annotations, which inevitably introduces discretisation noise.

To obtain a measurement that is robust to this scale mismatch, we additionally compute pairwise win-rate matrices for both judges, reporting the fraction of samples on which each row-model is preferred over each column-model (Table~\ref{tab:winrate}). Three observations emerge. First, both judges produce the \textit{same ranking of models by average win-rate}, with C-FDM as the top-performing system, followed by GoT, SmartEdit, and C-FT. Second, the two judges agree on the single best model per sample 57\% of the time, well above the 25\% chance baseline for a four-way comparison. Third, GPT-4o is a \textit{conservative lower bound} on the performance of C-FDM relative to the specialised baselines: against GoT, human annotators prefer C-FDM 54.2\% of the time, whereas GPT-4o prefers it only 44.4\% of the time. This directly refutes the hypothesis that GPT-4o artificially inflates our model's scores out of a preference for VLM-generated images; if anything, GPT-4o \textit{under}-rates C-FDM relative to human judgement, and the improvements reported above should be read as a conservative estimate of the true gains from bootstrapping FDP from IDP.

\begin{table}[t]
\centering
\small
\setlength{\tabcolsep}{4pt}
\begin{tabular}{lcccc}
\toprule
\textbf{Human judge} & SmartEdit & C-FT & C-FDM & GoT \\
\midrule
SmartEdit       & --    & 0.750 & 0.611 & 0.611 \\
C-FT            & 0.222 & --    & 0.486 & 0.361 \\
C-FDM  & 0.361 & 0.486 & --    & 0.542 \\
GoT             & 0.361 & 0.611 & 0.458 & --    \\
\midrule
\textbf{GPT-4o judge} & SmartEdit & C-FT & C-FDM & GoT \\
\midrule
SmartEdit       & --    & 0.736 & 0.639 & 0.542 \\
C-FT            & 0.236 & --    & 0.278 & 0.250 \\
C-FDM  & 0.333 & 0.694 & --    & 0.444 \\
GoT             & 0.431 & 0.722 & 0.556 & --    \\
\bottomrule
\end{tabular}
\caption{Pairwise win-rate matrices for the human judge (top) and the GPT-4o judge (bottom). Entry $(i, j)$ is the fraction of samples on which model $i$ is preferred over model $j$. Both judges yield the same ranking.}
\label{tab:winrate}
\end{table}

\paragraph{Ablation Study on Synthetic Trajectories.} 
To assess the importance of extra supervision from IDM-synthetic trajectories,
Table~\ref{tab:ablation-study} reports GPT-4o's scores for this ablation. We see performance drops on most datasets---particularly on Something and AG---when the additional training data from unlabelled videos is removed, highlighting the effectiveness of bootstrapping C-FDM with large-scale real-world data via IDM. An exception is the WhatsUp dataset, which focuses on specific actions within a fixed scene; in this case, training in an open-domain setting may not transfer effectively. 

\paragraph{Ablation Study on Loss Weighting.} 
Based on Table~\ref{tab:ablation-study}, we also observe consistent degradation when loss weighting is removed, demonstrating the benefit of explicitly incorporating the recognition model into visual next-token prediction. To better understand the effect of loss weighting, Table~\ref{tab:gpt4o-score-analysis} reports the average scores for two criteria used in the GPT-4o-as-a-judge evaluation separately: Editing Success (ES), which measures how well the model captures the intended action and performs the corresponding edit, and Minimal Editing (ME), which assesses whether the model introduces unnecessary modifications. The full distribution of GPT-4o scores is provided in Appendix~\ref{appendix:gpt4o-detailed-es-oe}. Our analysis reveals that the primary bottleneck for C-FDM remains its ability to reliably follow the instruction, as reflected by the fact that ES scores are significantly lower than ME scores. Loss weighting partly solves this problem, increasing the editing success and reducing copying behaviour, albeit at the cost of sometimes over-editing the source observation.

\begin{figure}[t]
    \centering
    \includegraphics[width=\linewidth]{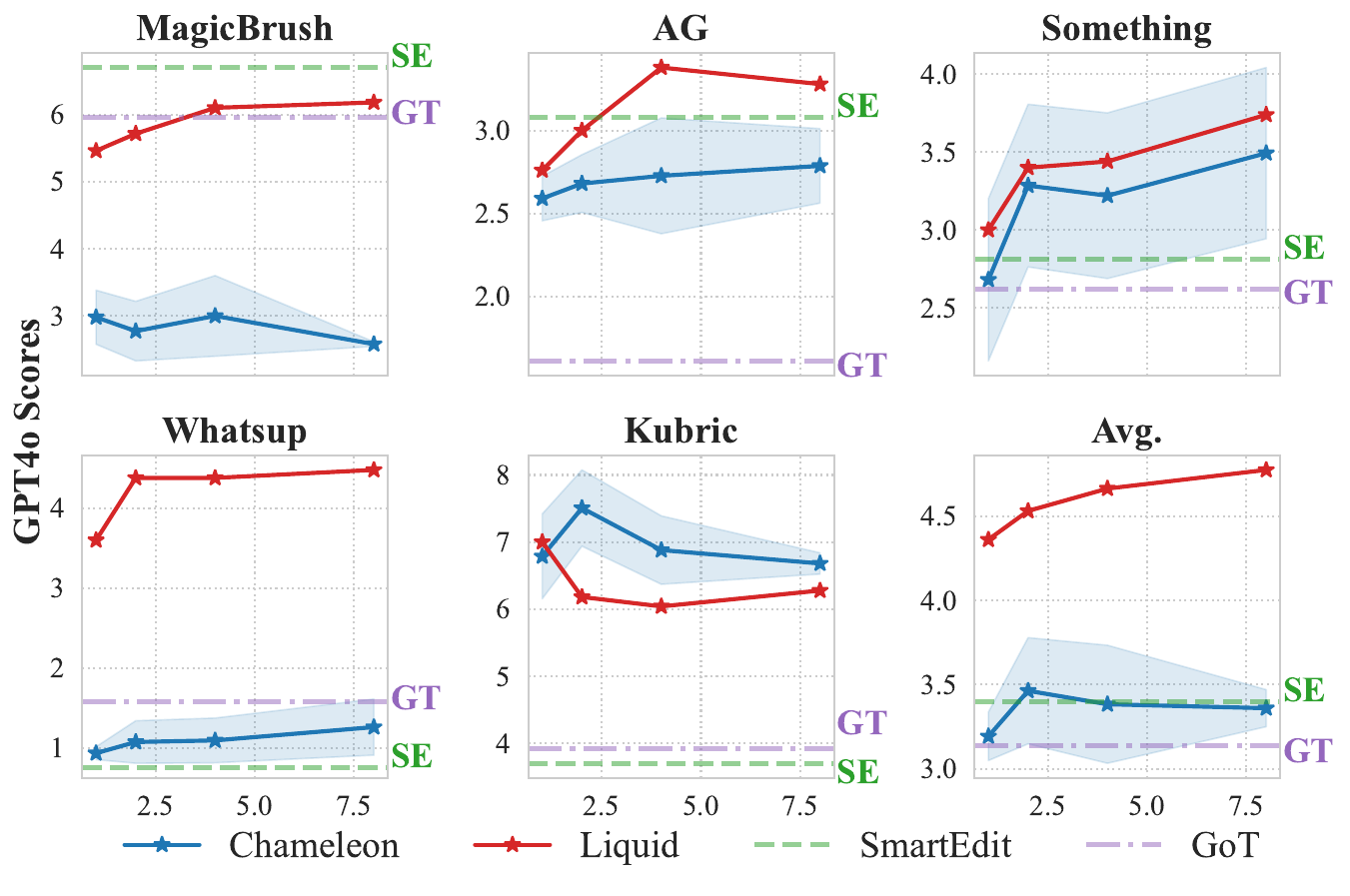}
    \caption{GPT-4o scores for test-time verification with $K$ samples, where $K \in \{1, 2, 4, 8\}$. We use a \textcolor{blue}{blue line} for C-FT and a \textcolor{red}{red line} for L-FT. For C-FT, we plot the standard deviation as the shaded area due to its large variance. We indicate the scores for GoT (GT) and SmartEdit (SE) as horizontal lines.}
    \label{fig:verification-at-k}
    
\end{figure}

\paragraph{Robustness of Loss-Weighting to Camera Motion.} One might worry that the recognition-weighted loss could be less effective under camera motion, where global pixel shifts could in principle wash out the token-level weighting signal. To test this empirically, we stratify \textsc{Aurora-Bench} by optical flow magnitude into a \textit{High Motion} subset (top 50\%) and a \textit{Low Motion} subset (bottom 50\%), and compare C-FDM against the C-FT baseline on each. Results are shown in Table~\ref{tab:motion-stratification}. Contrary to the concern, C-FDM delivers its \textit{largest} gains in high-motion scenarios: the improvement over C-FT is $+0.52$ GPT-4o points on the High Motion subset, compared to only $+0.11$ on the Low Motion subset. A plausible explanation is that in static scenes the baseline can exploit the copying heuristic discussed in our limitations, which preserves the source frame yields a competitive score with minimal risk, whereas in dynamic scenes this shortcut fails, and the loss-weighting objective successfully forces the model to attend to the semantically relevant regions of change. We therefore conclude that the recognition-weighted loss is not only robust to camera motion but is in fact most beneficial precisely in the high-motion regime where action-centric prediction is hardest.

\begin{table}[t]
\centering
\small
\begin{tabular}{lccc}
\toprule
& \textbf{C-FT} & \textbf{C-FDM} & $\Delta$ \\
\midrule
High Motion & 3.38 & 3.90 & +0.52 \\
Low Motion  & 3.40 & 3.50 & +0.11 \\
\bottomrule
\end{tabular}
\caption{GPT-4o-as-a-judge scores on \textsc{Aurora-Bench}, stratified by optical flow magnitude. 
The recognition-weighted loss yields its largest gains on high-motion samples.
}
\label{tab:motion-stratification}
\end{table}

\paragraph{Image Editing as an Auxiliary Task.}  
Finally, we assess whether enhanced FDP capabilities are beneficial for a broader range of vision--language tasks.
Since \textsc{Aurora} and the action-annotated videos contain diverse spatial relations (e.g., left/right orientation), we evaluated whether this supervision helps VLMs generalise beyond editing. We tested our models on two spatial reasoning benchmarks: SpatialMQA~\citep{liu2025spatialMQA} and EmbodiedSpatial-Bench~\citep{du2024embspatial}, and report the corresponding accuracy in Table~\ref{tab:main-result}. Both Chameleon and Liquid trained with the FDP objective outperform the zero-shot baseline, demonstrating that the FDP task transfers beyond action-centric editing and highlighting FDP as a signal for enhancing spatial reasoning.

\begin{figure*}[t]
    \centering
    \includegraphics[width=0.93\textwidth]{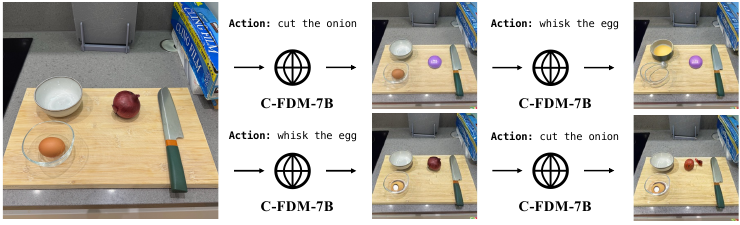}
    \caption{A qualitative case of real-world next-observation prediction, demonstrating C-FDM's ability to steer predictions using language and perform \textit{sequential} predictions. More cases from \textsc{Aurora-Bench} are in Appendix~\ref{appendix:qualitative-case}.}
    \label{fig:qualitative-example}
\end{figure*}

\subsubsection{Inference-time Verification} 
\label{sec:res_infver}
We evaluate our test-time strategy using IDM to choose among candidates generated by C/L-FT in Figure~\ref{fig:verification-at-k}, using $K \in {1, 2, 4, 8}$. 
By increasing exploration on more candidate next observations, Chameleon and Liquid benefit from test-time verification on most datasets with real-world trajectories (e.g., AG, Something, WhatsUp), indicating the reliability of IDM's trajectory preferences. 
Increasing $K$ does not always improve performance (MagicBrush, Kubric), however, suggesting that bootstrapping with IDM that shares the same foundation model backbone may be limiting. 
In summary, IDM-based verification boosts performance to a similar level as FDM, by leveraging more samples during inference rather than training.

\paragraph{Qualitative Example.} Figure~\ref{fig:qualitative-example} presents a real-world example demonstrating that C-FDM is also capable of iteratively generating future observations in multiple steps while maintaining consistency with previous frames.

\section{Related Work}

Despite the surge in interest for world modelling \citep{2018_ha_schmidhuber_recurrent_world_model,sutton1988worldmodeling,hafner2019learningLatentDynamics}, previous works focused mostly on building specialised \textit{ad-hoc} world models. These world models can be explicitly learnt as a visual simulator \citep{agarwal2025cosmos,bruce2024genie,brooks2024videoGenerationWorldSimulator}, or enable planning with model-based reinforcement learning \citep{hafner2019learning,micheli2022transformers,robine2023transformerworldmodelTWM,alonso2024diffusionForWorldModeling,Hafner2025}. Instead, we focus on leveraging general-purpose, pre-trained vision-language models \citep{lin2024video,chen2025janus,wu2024vilaU} to develop world models, which is more attractive due to the inductive bias they provide from their extensive training. 

This is possible thanks to frameworks that integrate observations, actions, and rewards into a unified sequence of tokens in autoregressive Transformers \citep{NEURIPS2024_7dbb5bfa}, building on pioneering works such as Decision Transformers  \citep{chen2021decisionTransformers} and GATo \citep{reed2022generalistGATo}. Related to our work, \citet{chen-vlm-representation-for-RL} initialise the parameters of RL policies with VLMs, thus taking advantage of the abundant and general world knowledge encoded in their representations. 3D-VLA \citep{zhen20243dWorldModelVLM} integrates a set of interaction tokens into a Large Language Model to engage with the environment as an embodied agent. %
\citet{yang2024video,qiu2026swirl,soni2024videoagent} explore large-scale self-supervised learning via next token or frame prediction to build a unified model absorbing internet knowledge, learning from interaction via video.

Most similar to our work, \citet{baker2022videoPretraining-vpt} train a dynamics model which aims to uncover the underlying action between video frames in unlabelled video frames from the Minecraft game. Through this model, they synthesise trajectories to train a policy for sequential decision making. In contrast with \citet{baker2022videoPretraining-vpt}, we focus on actin-centric next-observation prediction as a task to evaluate FDP. First, this allows us to port the observation space to real-world frames, rather than simulated ones, hence assessing whether VLMs can eventually be developed into world models. Second, this broadens the space of actions from a few choices to the combinatorially infinite and expressive space of language, capturing a significantly more diverse range of dynamics.

\section{Conclusions}
\label{sec:conclusion}

In this work, we explored whether unified vision--language models (VLMs) can be endowed with the ability to predict forward dynamics, i.e., the next observation in the environment (e.g., an image frame) given the past observation and an action (e.g., a textual instruction). We first show that these models lack a clear preference for ground-truth real-world action--observation trajectories compared with adversarially manipulated ones. 

To address this, we leverage an inverse-dynamics model (IDM) fine-tuned from the same VLM, which consists instead of predicting actions taking place between observations and is easier to learn, to bootstrap a better forward-dynamics model (FDM). Specifically, the IDM can be used to 1) automatically annotate pairs of frames from unlabelled videos, which are then used for weakly supervised training of the FDM; or 2) verify the best sample among multiple candidates generated from the FDM at inference time. Experiments confirm the effectiveness of both strategies, with our general-purpose forward-dynamics model achieving state-of-the-art performance compared to existing approaches specialised for image editing.

\section*{Limitations}
\label{appendix:limitations}

While overall our results demonstrate the effectiveness of our approaches across \textsc{Aurora-Bench}, we would like to highlight a few limitations that we have discovered: 
\begin{itemize}
    \item Despite efforts to guide the model via weakly supervised fine-tuning with loss weighting or inference-time verification (Table~\ref{tab:main-result}), we observe that the model may still resort to copying the source observation, especially under low sampling temperatures or ambiguous instructions.
    \item While we show promising preliminary results of language-steered next-observation prediction in Figure~\ref{fig:qualitative-example}, fine-grained control remains limited, and understanding subtle instructions (e.g., spatial or quantitative edits) remains challenging.
    \item We observe high variance across different runs of experiments for Chameleon, likely due to the sensitivity of sampling from a weak model. To address this, we report results averaged over multiple runs. %
\end{itemize}

\section*{Acknowledgements}

We thank the reviewers and the area chair for their valuable comments.
Yifu Qiu is grateful for an Apple AI/ML scholarship that supported this work. This work is supported by the ERC Starting Grant AToM-FM (101222956) awarded to Edoardo M. Ponti. We are thankful for the compute resources provided
by UKRI through the Isambard AI cluster (managed by the University of Bristol).

\bibliography{custom}

\newpage

\appendix

\section{Potential Risks}
\label{appendix:boarder-impact}

This work develops models for action-centric image editing for visual world modelling. While our primary aim is to advance fundamental research in world modelling, we acknowledge potential risks, particularly in the generation of realistic future observations.

A core concern is the potential misuse of the models for creating deceptive visual content, including fabricated action sequences or manipulated images that imply false causality. Although the model is not explicitly designed for these tasks, its ability to generate coherent visual predictions from the linguistic action could be adapted for such uses if deployed irresponsibly.

Even in intended use, risks include over-reliance on generated outputs in downstream tasks such as robotic control, or interactive systems. Model failures—e.g., copying artefacts, hallucinations, or broken object continuity—can lead to incorrect inferences or reinforce dataset biases.

To mitigate potential misuse, we limit our model release to research purposes under a non-commercial license and clearly communicate its capabilities and limitations. We urge caution when adapting them for deployment, particularly in settings with high societal or ethical sensitivity.

\section{Model Performance on \textsc{Aurora-Bench} with 5 Metrics}
\label{appendix:all-5-metrics-results}

In addition to GPT4o-as-a-judge evaluation, we further employ a diverse set of automatic metrics covering both low-level and semantic fidelity: 1) we compute the \textbf{L1 distance} between the predicted and target observation as a pixel-level metric. 2) We extract visual features and compute the cosine similarity in their respective embedding spaces for several image encoders, including (\textbf{CLIP-I} and \textbf{DINO}), to assess semantic similarity. Additionally, to measure alignment between image content and the action semantics, we compute \textbf{CLIP-T}, the similarity between the edited image and its BLIP-generated caption. These metrics are evaluated in addition to GPT4o-as-a-judge metric following previous works in image editing \citep{huang2024smartedit,fang2025got,krojer2024aurora}. We report the detailed results with 5 metrics in Table~\ref{tab:main-result-5-metrics}. We notice that copy baseline exhibits the best performance as measured by the distance-based and visual encoder-based approach, as indicated in Table~\ref{tab:main-result}. This poses a challenge to the reliability of the traditional metrics in fairly evaluating the action-centric image editing task. On the other hand, GPT4o-as-a-judge metric robustly assigns $0$ score to Copy, indicating its robustness in detecting copying generation while putting GPT-as-a-judge as the most reliable metric to interpret.

\begin{table*}[h]

\resizebox{\textwidth}{!}{
\begin{tabular}{ccccccccccc}
\toprule
\multirow{2}{*}{\textbf{Datasets}}   & \multirow{2}{*}{\textbf{Metrics}} & \multicolumn{9}{c}{\textbf{Models}}                                                                                                                                     \\ \cmidrule{3-11} 
                                     &                                   & \textbf{Copy} & \textbf{PixInstruct} & \textbf{GoT}   & \textbf{SE} & \textbf{CM} & \textbf{\small{C-FT}} & \textbf{\textit{+Best-of-3}} & \textbf{C-FDM} & \textbf{\textit{+Best-of-3}} \\ \midrule
\multirow{5}{*}{\textbf{MagicBrush}} & L1                                & 0.027         & 0.114                & \textbf{0.063} & 0.068              & 0.287              & 0.075               & 0.075        & 0.090             & 0.078        \\
                                     & CLIP-I                            & 0.959         & 0.877                & 0.930          & \textbf{0.937}     & 0.671              & 0.913               & 0.914        & 0.906             & 0.909        \\
                                     & CLIP-T                            & 0.289         & 0.275                & 0.286          & 0.290              & 0.227              & 0.289               & 0.289        & \textbf{0.291}    & 0.291        \\
                                     & DINO                              & 0.931         & 0.761                & 0.881          & \textbf{0.894}     & 0.292              & 0.883               & 0.883        & 0.864             & 0.864        \\
                                     & GPT-4o                            & 0.000         & 3.120                & 5.960          & \textbf{6.710}     & 0.000              & 2.520               & 3.270        & \textcolor{myblue}{3.920}             & \textcolor{myblue}{3.920}        \\ \midrule
\multirow{5}{*}{\textbf{AG}}         & L1                                & 0.069         & 0.220                & 0.174          & \textbf{0.137}     & 0.314              & 0.170               & 0.168        & 0.168             & 0.167        \\
                                     & CLIP-I                            & 0.943         & 0.757                & 0.846          & 0.811              & 0.609              & 0.872               & 0.872        & \textbf{0.881}    & 0.883        \\
                                     & CLIP-T                            & 0.279         & 0.254                & 0.280          & 0.268              & 0.214              & 0.280               & 0.284        & \textbf{0.284}    & 0.284        \\
                                     & DINO                              & 0.929         & 0.557                & 0.785          & 0.774              & 0.258              & 0.801               & 0.817        & \textbf{0.816}    & 0.816        \\
                                     & GPT-4o                            & 0.000         & 1.200                & 1.610          & 3.080              & 0.170              & 2.480               & 2.740        & \textcolor{myblue}{\textbf{3.640}}    & \textcolor{myblue}{3.640}        \\ \midrule
\multirow{5}{*}{\textbf{Something}}  & L1                                & 0.135         & 0.232                & 0.184          & \textbf{0.163}     & 0.293              & 0.184               & 0.184        & 0.196             & 0.184        \\
                                     & CLIP-I                            & 0.870         & 0.709                & 0.807          & 0.773              & 0.649              & \textbf{0.820}      & 0.820        & 0.804             & 0.804        \\
                                     & CLIP-T                            & 0.275         & 0.238                & 0.269          & 0.265              & 0.232              & \textbf{0.271}      & 0.269        & 0.268             & 0.268        \\
                                     & DINO                              & 0.797         & 0.453                & 0.636          & 0.662              & 0.297              & \textbf{0.675}      & 0.653        & 0.666             & 0.666        \\
                                     & GPT-4o                            & 0.000         & 0.957                & 2.620          & 2.810              & 0.370              & \textcolor{myblue}{\textbf{3.110}}      & 3.110        & 2.920             & \textcolor{myblue}{3.310}        \\ \midrule
\multirow{5}{*}{\textbf{WhatsUp}}    & L1                                & 0.039         & 0.138                & 0.078          & 0.067              & 0.251              & \textbf{0.066}      & 0.066        & 0.070             & 0.070        \\
                                     & CLIP-I                            & 0.954         & 0.817                & \textbf{0.923} & 0.888              & 0.721              & 0.877               & 0.880        & 0.870             & 0.883        \\
                                     & CLIP-T                            & 0.326         & 0.287                & \textbf{0.316} & 0.312              & 0.243              & 0.309               & 0.310        & 0.306             & 0.307        \\
                                     & DINO                              & 0.908         & 0.615                & \textbf{0.850} & 0.805              & 0.424              & 0.836               & 0.841        & 0.831             & 0.838        \\
                                     & GPT-4o                            & 0.000         & 0.000                & \textbf{1.580} & 0.755              & 0.146              & \textcolor{myblue}{0.880}               & \textcolor{myblue}{0.980}        & 0.540             & 0.540        \\ \midrule
\multirow{5}{*}{\textbf{Kubric}}     & L1                                & 0.011         & 0.104                & \textbf{0.026} & 0.064              & 0.276              & 0.044               & 0.044        & 0.044             & 0.044        \\
                                     & CLIP-I                            & 0.963         & 0.796                & 0.895          & 0.868              & 0.660              & 0.897               & 0.899        & \textbf{0.897}    & 0.898        \\
                                     & CLIP-T                            & 0.282         & 0.259                & 0.281          & 0.271              & 0.213              & \textbf{0.287}      & 0.287        & 0.287             & 0.288        \\
                                     & DINO                              & 0.955         & 0.676                & 0.857          & 0.798              & 0.161              & \textbf{0.906}      & 0.906        & 0.902             & 0.902        \\
                                     & GPT-4o                            & 0.000         & 1.880                & 3.920          & 3.700              & 0.140              & 7.300               & 7.300        & \textcolor{myblue}{\textbf{7.320}}    & \textcolor{myblue}{7.780}        \\ \midrule
\textbf{All}                        & GPT-4o                            & 0.000         & 1.430                & 3.140          & 3.410              & 0.165              & 3.260               & 3.480        & \textcolor{myblue}{\textbf{3.670}}    & \textcolor{myblue}{3.840}        \\ \bottomrule
\end{tabular}
}
\caption{Model performance at MagicBrush, Action-Genome, Something, WhatsUp and Kubric on \textsc{Aurora-Bench}. For C-FT and C-FDM We report both the model performance and their performance in the \textit{best-of-N} distribution. We report the average GPT4o scores for each model at the bottom. 
We \textcolor{myblue}{highlight} the better GPT-4o scores for C-FT and C-FDM.
We \textbf{bold} the best performance among all models, except Copy and \textit{best-of-N} performances.
SE: SmartEdit.}
\label{tab:main-result-5-metrics}
\end{table*}

\begin{figure*}[t]  %
    \centering
    \includegraphics[width=\linewidth]{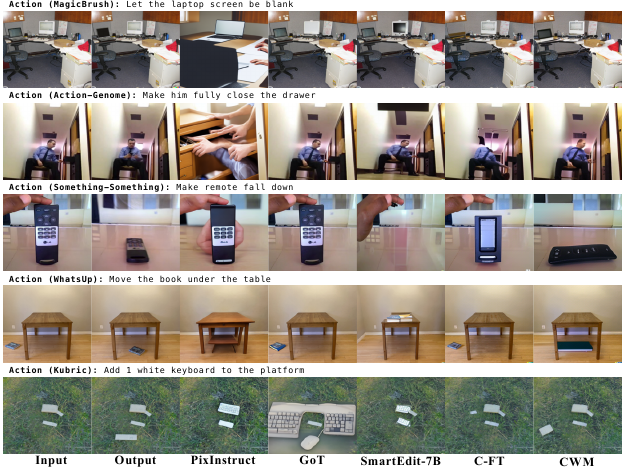}
    \caption{Qualitative examples of the predicted next observation from the state-of-the-art specialised image editing models, and our models including C-FT and C-FDM, on \textsc{Aurora-Bench}.}
    \label{fig:aurora-qualitative-case}
\end{figure*}

\section{Qualitative Cases}
\label{appendix:qualitative-case}

In this section, we present additional qualitative examples from \textsc{Aurora-Bench} in Figure~\ref{fig:aurora-qualitative-case}. We observe several common failure modes in image editing models. First, they sometimes fail to preserve the scene from the source observation (e.g., PixInstruct on Action-Genome and MagicBrush). Second, some models generate near-identical copies of the source as the target (e.g., GoT on Something-Something). Third, producing realistic outputs remains difficult, as seen in GoT’s result on Kubric. Finally, maintaining object consistency is also a challenge—SmartEdit alters the object in WhatsUp, and C-FDM does so in Something-Something.

Despite the challenges, we also observe several positive editing behaviours from C-FDM. On Action-Genome, C-FDM correctly predicts spatial changes, such as \textit{opening and closing a drawer}, which requires a strong understanding of the spatial concepts. In Something-Something, it is the only model to accurately capture the spatial concept of “falling down.” On Kubric, it demonstrates basic counting ability by correctly adding one keyboard. In WhatsUp, C-FDM correctly grounds the action to the laptop, while other models mistakenly edit the monitor.

\section{Detailed Discussion for Chameleon's Predicted Likelihoods}
\label{appendix:detailed-likelihood-anlaysis}

\begin{figure*}[t]
    \centering
    \begin{minipage}{\textwidth}
        \centering
        \includegraphics[width=\textwidth]{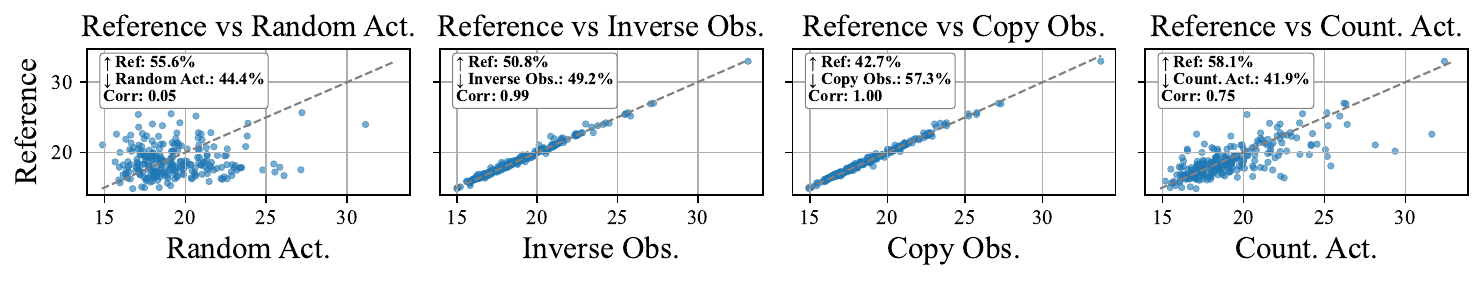}
    \end{minipage}
    
    \begin{minipage}{\textwidth}
        \centering
        \includegraphics[width=\textwidth]{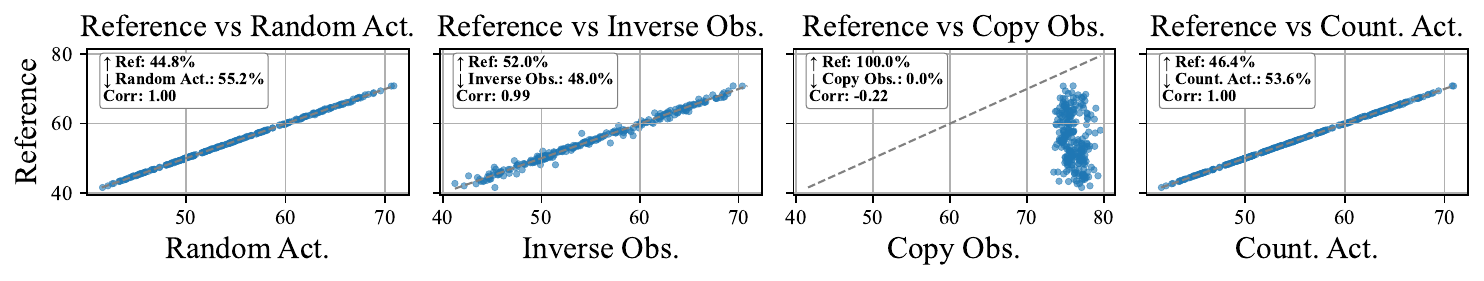}
    \end{minipage}
    
      \caption{
      Comparison of predicted negative log-likelihoods (lower values indicate stronger model preference) for ground-truth real-world trajectories versus four types of negative trajectories.
        \textbf{Top}: Action prediction task for the IDP (observation $\times$ observation $\rightarrow$ action).
        \textbf{Bottom}: Next observation prediction task for the FDP (observation $\times$ action $\rightarrow$ observation).
        The legend shows the percentage of times the model prefers the ground-truth trajectory ($\uparrow$) over the negatives ($\downarrow$).
    }

  \label{fig:chameleson-zs-log-likelihood-comparison}
\end{figure*}

\begin{figure*}
    \centering
    \includegraphics[width=\linewidth]{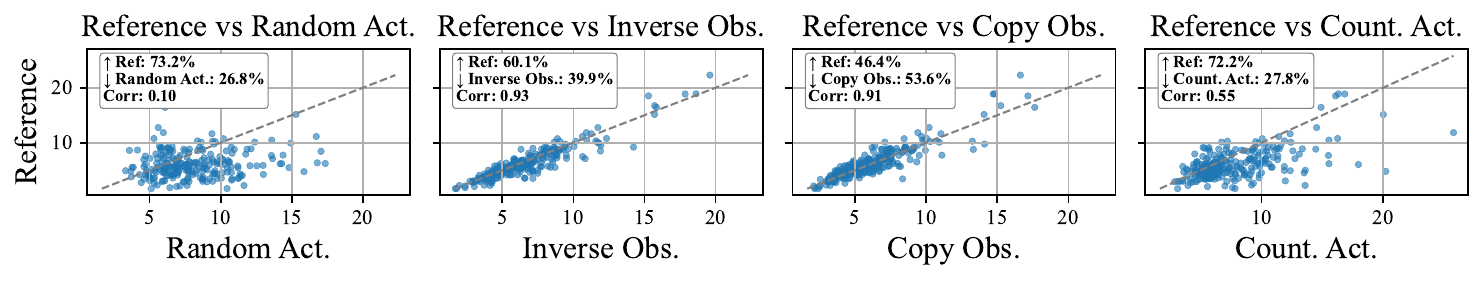}
  
  \caption{
      Comparison of negative log-likelihoods (lower values indicate stronger model preference) of the action predicted by IDM for ground-truth trajectories versus four types of negative trajectories.
  }
  
  \label{fig:action-prediction-min-loss-percentage}
\end{figure*}

From Figure~\ref{fig:chameleson-zs-log-likelihood-comparison}, it emerges that Chameleon-7B displays a very limited preference for the ground-truth trajectories in a zero-shot setting. In the action prediction task (top panel), there is a slightly higher tendency to favour the ground-truth; however, even in the best case (counterfactual action), the model prefers the reference in only 58.1\% of the samples. The high correlation in likelihoods indicates that the VLM struggles also on visual manipulations.
In the next-observation prediction task (bottom panel), the VLM mostly fails in effectively differentiating the ground truth from the negatives. An exception to this is the copy manipulation, where the model can always tell them apart. Although the underlying reason remains uncertain, one plausible explanation for this behaviour is that the model's ability to solve next-observation prediction tasks depends on their alignment with training sequences: for instance, it is plausible that Chameleon's data rarely features two identical adjacent images. 
In Figure~\ref{fig:action-prediction-min-loss-percentage}, we visualize the predicted likelihoods produced by Chameleon’s fine-tuned inverse dynamics model (IDM). We observe that fine-tuning on ground-truth trajectories substantially increases the model’s ability to distinguish ground-truth actions from negative alternatives. Specifically, the probability that the ground-truth action is assigned a higher likelihood increases from 55.6\% to 73.2\% under random-action negatives, and from 58.1\% to 72.2\% under counterfactual-action negatives. These results demonstrate the strong potential of learning effective inverse dynamics models directly from real-world trajectories.
In summary, the zero-shot Chameleon-7B does not exhibit a preference for ground-truth trajectories over negative ones, constructed through action- or observation-based manipulations. However, it is possible to learn an effective IDM from the real-world trajectories.

\section{Details of Processing IDM Annotations for Unlabelled Videos}
\label{appendix:stratified-top-k}

\begin{figure*}[p]
    
    \rotatebox{-90}{%
        \begin{minipage}{\textheight}
            \centering
            \begin{subfigure}{0.8\textwidth}
                \centering
                \includegraphics[width=\textwidth]{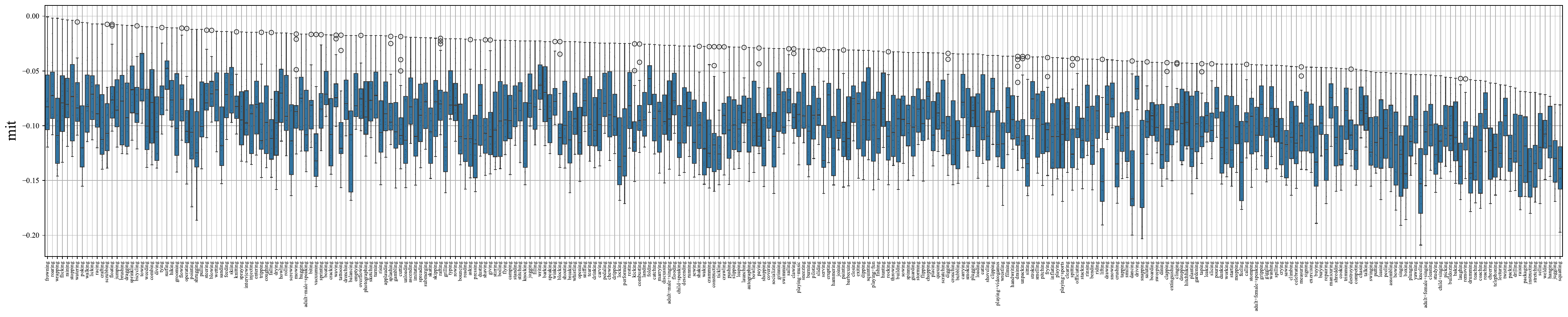}
            \end{subfigure}

            \begin{subfigure}{0.8\textwidth}
                \centering
                \includegraphics[width=\textwidth]{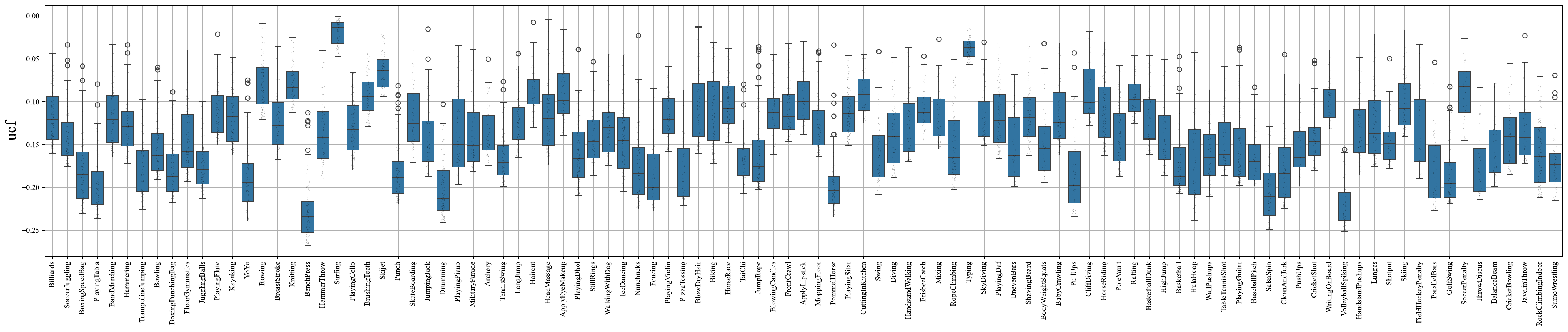}
            \end{subfigure}

            \begin{subfigure}{0.8\textwidth}
                \centering
                \includegraphics[width=\textwidth]{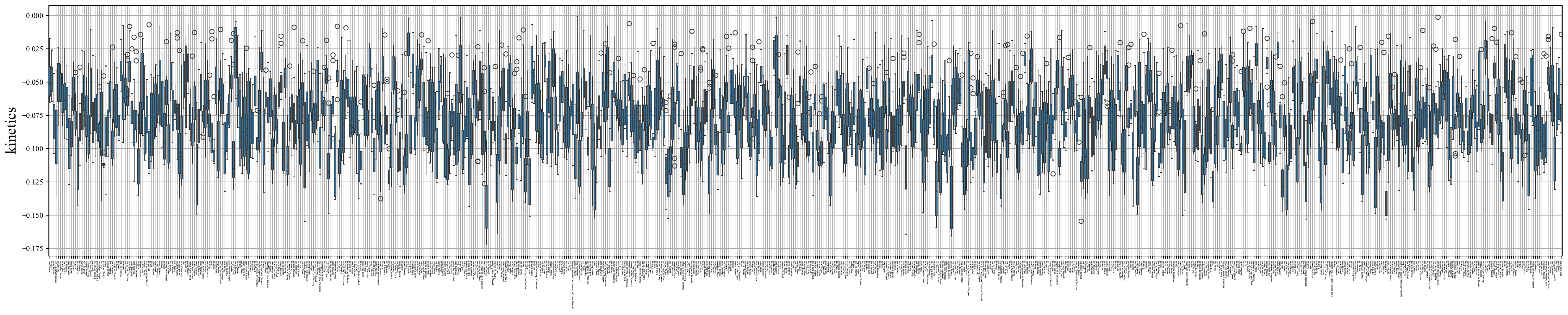}
            \end{subfigure}
        \end{minipage}
    }   
        \caption{Distributions of triplet log-likelihoods predicted by IDM on Movements-in-Time, UCF-101, and Kinetics-700, based on 7K synthetic triplets per dataset. Triplets are uniformly sampled from each action class while maximising overall predicted likelihoods.}
        \label{fig:scores-per-action-class}
    
\end{figure*}

\begin{algorithm*}[t]
\caption{Stratified Top-K Sampling with Action Class Uniformity}
\begin{algorithmic}[1]
\Require Trajectory triplet set $X = \{(o_s^i, o_t^i, a^i, s^i, c^i)\}_{i=1}^{N}$, where $s_i$ is the predicted likelihood of $a^i$, $c_i \in \mathcal{C}$ is the class, number of samples $K$

\State Sort $X$ descending by score $s_i$
\State Initialize $S \gets \emptyset$, and $\text{class\_counts}[c] \gets 0$ for all $c \in \mathcal{C}$

\While{$|S| < K$}
    \ForAll{class $c \in \mathcal{C}$ in round-robin order}
        \State $X_c \gets$ top unsampled item from class $c$ in $X$
        \If{$X_c \neq \emptyset$}
            \State $S \gets S \cup \{X_c\}$
            \State Remove $X_c$ from $X$
            \State $\text{class\_counts}[c] \gets \text{class\_counts}[c] + 1$        \EndIf
        \If{$|S| = K$}
            \State \textbf{break}
        \EndIf
    \EndFor
\EndWhile
\State \Return $S$
\end{algorithmic}
\label{algo:IDM-annotation-sampling}
\end{algorithm*}

We present the raw dataset statistics before sampling for Movements-in-Time, UCF-101 and Kinetics700 in Table~\ref{tab:IDM-trajectories-stat}.
Figure~\ref{fig:scores-per-action-class} shows the distribution of IDM's predicted scores across action classes in Movements-in-Time, Kinetics700, and UCF-101. The predicted likelihoods are nearly uniform within each class, indicating that our sampling method maintains both class diversity and high overall likelihoods. The sampling procedure for IDM-annotated trajectories is detailed in Algorithm~\ref{algo:IDM-annotation-sampling}.

\begin{table*}[]
\resizebox{\textwidth}{!}{
\begin{tabular}{lcc|cccc}
\toprule
\multicolumn{1}{c}{\multirow{2}{*}{\textbf{Dataset}}} & \multicolumn{2}{c|}{\textbf{Video}}           & \multicolumn{4}{c}{\textbf{Triplet}}                                                 \\ \cmidrule{2-7} 
\multicolumn{1}{c}{}                                  & \textbf{Avg. Length} & \textbf{Total Length} & \textbf{\#Samples} & \textbf{\#Avg. OPV} & \textbf{\#Avg. APV} & \textbf{\#Avg. WPA} \\ \midrule
\textbf{MIT}                                          & 3.04 seconds         & 2.57 hours            & 19,658              & 2.05                & 1.05                & 7.10                \\
\textbf{UCF-101}                                      & 7.24 seconds         & 26 hours              & 10,965              & 3.00                & 2.00                & 8.96                \\
\textbf{Kinetics700}                                  & 9.02 seconds         & 18 hours              & 26,959              & 2.71                & 1.71                & 7.39                \\ \bottomrule
\end{tabular}
}
\caption{Dataset statistics for the video and triplets from the trajectories annotated by IDM. \textbf{OPV}: observations (i.e., extracted key-frames) per video, \textbf{APV}: actions per video, \textbf{WPA}: words per action.}

\label{tab:IDM-trajectories-stat}
\end{table*}

\section{Prompt Template for Using GPT4o-as-a-Judge}
\label{appendix:gpt4o-judge}

\begin{figure*}[t]
\centering
\begin{tcolorbox}[
    title=Prompt Template for GPT4o-as-a-judge Evaluation,
    width=\textwidth,
    colback=gray!5,
    colframe=gray!80
]

You are a professional digital artist. You will have to evaluate the effectiveness of the AI-generated image(s) based on the given rules. \\

You will have to give your output in a valid way of a Python dictionary format (Keep your reasoning concise and short.): \\

\texttt{\{\{"score": [...], "reasoning": "..." \}\}} \\

and don’t output anything else. Two images will be provided:

\begin{itemize}
    \item The first being the original AI-generated image
    \item The second being an edited version of the first.
\end{itemize}

The objective is to evaluate how successfully the editing instruction has been executed in the second image.

Note that sometimes the two images might look identical due to a failure in image editing. From a scale of 0 to 10:

\begin{itemize}
    \item A score from 0 to 10 will be given based on the success of the editing.
    \item A second score from 0 to 10 will rate the degree of minimal editing.
\end{itemize}

Editing instruction: \{instruction\}

\end{tcolorbox}
\caption{Prompt template used for GPT-4o-as-a-judge evaluation.}
\label{fig:gpt4o-prompt}
\end{figure*}

We provide the prompts used for evaluating image editing performance with GPT-4o in Figure~\ref{fig:gpt4o-prompt}. We use \texttt{GPT-4o-2024-11-20}. The final score is the average of the minimum value of the two scores for each sample, as in \citep{fang2025got}.

\newpage

\section{Detailed GPT4o Scores for Editing Success and Minimal Editing}
\label{appendix:gpt4o-detailed-es-oe}

\begin{figure*}[htbp]
    \centering
    \begin{subfigure}[b]{\textwidth}
        \includegraphics[width=\textwidth]{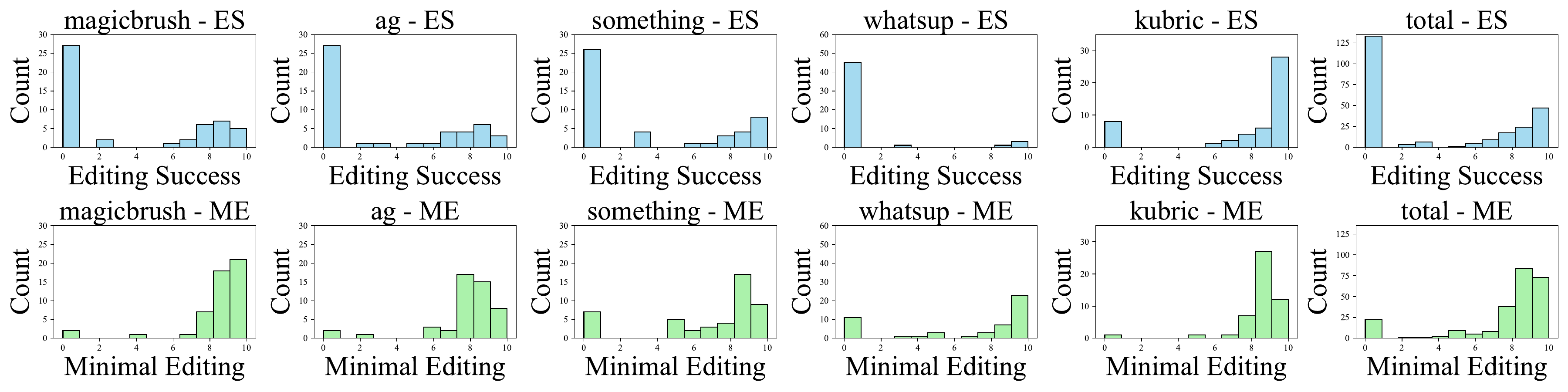}
        \caption{Detailed GPT4o scores for C-FDM trained with the standard loss.}
        \label{fig:sub1}
    \end{subfigure}

    \begin{subfigure}[b]{\textwidth}
        \includegraphics[width=\textwidth]{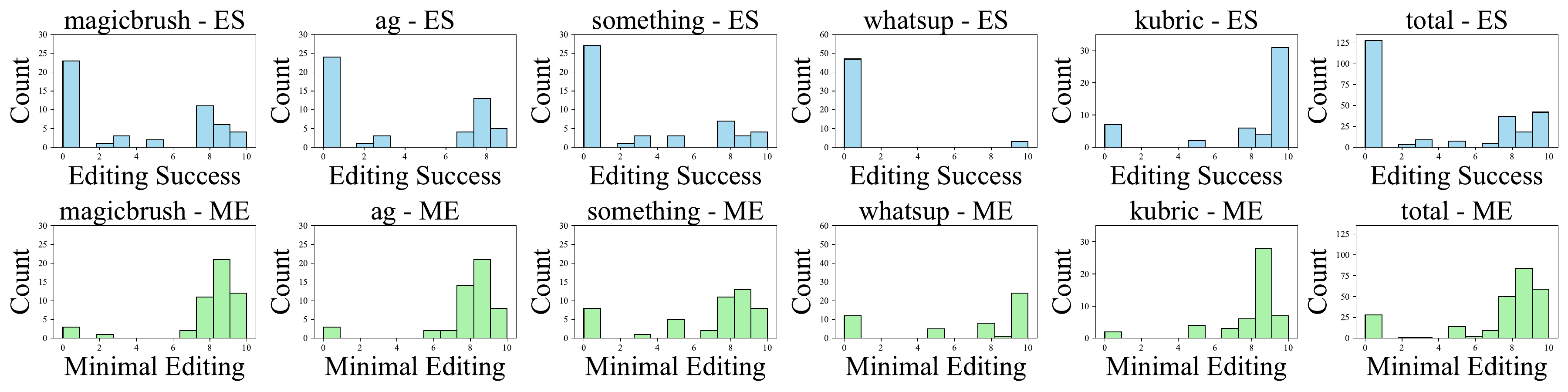}
        \caption{Detailed GPT4o scores for C-FDM trained with the L$_2$-weighted loss.}
        \label{fig:sub2}
    \end{subfigure}
    \caption{GPT4o scores' distributions of editing success (ES) and minimal editing (OE) for C-FDM trained with standard loss or our loss-weighting method. }
    \label{fig:gpt4o-detailed-score}
\end{figure*}

Figure~\ref{fig:gpt4o-detailed-score} shows the distribution of editing success (ES) and minimal editing (ME) scores for standard training and loss-weighted training. Loss weighting tends to improve editing success, with a modest trade-off in minimal editing quality in most of the datasets.

\section{Implementation Details}
\label{appendix:implementation-details}

\subsection{Chameleon Dynamics Model}
\label{appendix:training-details-IDM}
We fine-tune the Chameleon-7B checkpoint from the Anole-7B version \citep{chern2024anole} to predict the action given a pair of observations, framed as an action-prediction task. The model is trained on a merged dataset from Action-Genome, Kubric, MagicBrush, Something-Something from AURORA's annotated trajectories, and 15K EPIC-Kitchens processed by us. We downsample Kubric's trajectories to 10K. Training is performed for 10 epochs with a batch size of 64, using a learning rate of 2e-4 and cosine scheduling (500 warm-up steps). We use bfloat16 mixed-precision training and apply LoRA \citep{hulora} for parameter-efficient fine-tuning (rank 16, $\alpha=32$, dropout 0.05). Only the completion loss is used to optimise the generation of action. Training is conducted on 4 NVIDIA-H100-80GB-HBM3 GPUs using DeepSpeed for distributed optimisation.

\subsection{C-FT Baseline}

We fine-tune the Chameleon-7B checkpoint from the Anole-7B version \citep{chern2024anole}. The model is trained on a combined dataset from Action-Genome, Kubric, MagicBrush, and Something-Something, formatted as the image editing task. We downsample Kubric's trajectories to 10K. Training is conducted for 40 epochs with a batch size of 96 using the AdamW optimiser and a cosine learning rate scheduler (learning rate of 5e-4, 400 warm-up steps). We use mixed-precision training with bfloat16 and apply LoRA \citep{hulora} for efficient fine-tuning (rank 16, $\alpha=32$, dropout 0.05). We only train the model with the truncated loss from the completion part. We use 4 NVIDIA-H100-80GB-HBM3 GPUs with DeepSpeed for distributed training. During inference, we apply a logits processor to mask out non-image tokens, set the temperature to 1, and use top-1 sampling. We observe that temperature is critical in controlling model behaviour: lower values often cause the model to copy the source observation instead of generating meaningful edits.

\subsection{Chameleon FDM}

We fine-tune the Chameleon-7B checkpoint from the Anole-7B version \citep{chern2024anole}. The model is trained on a combined dataset from Action-Genome, Kubric, MagicBrush, Something-Something from AURORA's annotated trajectories, together with 7K trajectories from Movements-in-Time, 7K trajectories from UCF-101 and 7K trajectories from Kinetics700, formatted as the image editing task. Again, we downsample Kubric's trajectories to 10K. Training is conducted for 40 epochs with a batch size of 96 using the AdamW optimiser and a cosine learning rate scheduler (learning rate of 5e-4, 400 warm-up steps). We use mixed-precision training with bfloat16 and apply LoRA \citep{hulora} for efficient fine-tuning (rank 16, $\alpha=32$, dropout 0.05). We only train the model with the truncated loss from the completion part, but we weight the image tokens using L$_2$ strategy as introduced in Section~\ref{sec:methodology}. We use 4 NVIDIA-H100-80GB-HBM3 GPUs with DeepSpeed for distributed training. We use the same hyperparameters as C-FT during the inference time.

\subsection{Computing Resources}
\label{appendix:coputing-resources}

All training experiments were conducted on a compute node equipped with 4× NVIDIA H100 80GB GPUs, 256 CPU cores, and 256GB of memory. The total GPU hours required for training C-FT, C-FDM, and IDM were approximately 200, 400, and 100 hours, respectively.

For inference, we used a single NVIDIA A100 80GB GPU with 8 CPU cores and 128GB memory. Inference for C-FT and C-FDM takes approximately 1 GPU hour per model. When applying verification with $K=8$, inference time increases to around 8 GPU hours. IDM only takes around 0.3 GPU hours for inference.

\subsection{Assets and Licenses}
\label{appendix:assets-license}

In this section, we list the public assets we used in this paper and the corresponding links.

\noindent \textbf{Datasets.} We include the detailed license and URL for the datasets we used in this paper. 
\begin{itemize}
    \item \textsc{Aurora} and \textsc{Aurora-Bench} \citep{krojer2024aurora}: MIT license, the reader can find the corresponding version we use in this paper in \url{https://github.com/McGill-NLP/AURORA}.
    \item Movements-in-Time \citep{monfortmoments-MIT}: BSD-2-Clause license and its own License for Non-Commercial Use, the reader can find the corresponding version we use in this paper in \url{http://moments.csail.mit.edu/}.
    \item UCF-101 \citep{soomro2012ucf101}: unknown license, the reader can find the corresponding version we use in this paper in \url{https://huggingface.co/datasets/flwrlabs/ucf101}.
    \item Kinetics700 \citep{kinetics700, carreira2019kinetics700Note}: Creative Commons Attribution 4.0 International License, the reader can find the corresponding version we use in this paper in \url{https://research.google/pubs/the-kinetics-human-action-video-dataset/}.
    \item EPIC-Kitchens \citep{damen2018epic-kitchens}: Creative Commons Attribution-NonCommercial 4.0 International License, the reader can find the corresponding version we use in this paper in \url{https://epic-kitchens.github.io/}.
\end{itemize}

\noindent \textbf{Implementation.} We use the other following code for the implementations:

\begin{itemize}
    \item Transformers \citep{wolf-etal-2020-transformers}: Apache-2.0 license. We use the 4.47.0 version, following the link at \url{https://github.com/huggingface/transformers}.
    \item DeepSpeed: We use the 0.14.4 version, following the link at \url{https://github.com/deepspeedai/DeepSpeed}.
\end{itemize}

\noindent \textbf{Model.} We use the following models or checkpoints for the implementations:

\begin{itemize}
    \item Chameleon \citep{team2024chameleon}: Chameleon Research License, the reader can find the corresponding version we use in this paper in \url{https://github.com/facebookresearch/chameleon}.
    \item Anole-7B \citep{chern2024anole}: Chameleon Research License and MIT License, the reader can find the corresponding version we use in this paper in \url{https://github.com/GAIR-NLP/anole}.
    \item VILA-U \citep{chern2024anole}: MIT License, the reader can find the corresponding version we use in this paper in \url{https://github.com/mit-han-lab/vila-u}.
    \item SmartEdit \citep{huang2024smartedit}: Apache-2.0, the reader can find the corresponding version we use in this paper in \url{https://huggingface.co/TencentARC/SmartEdit-7B}.
    \item GoT \citep{fang2025got}: MIT License, the reader can find the corresponding version we use in this paper in \url{https://github.com/rongyaofang/GoT}.
    \item PixInstruct \citep{brooks2023instructpix2pix}: PixInstruct customised license, the reader can find the corresponding version we use in this paper in \url{https://github.com/timothybrooks/instruct-pix2pix}.
\end{itemize}

\section{Details of Human Evaluation}
\label{appendix:human-eval-interface}

\begin{figure*}[t]
    \centering
    \includegraphics[width=\linewidth]{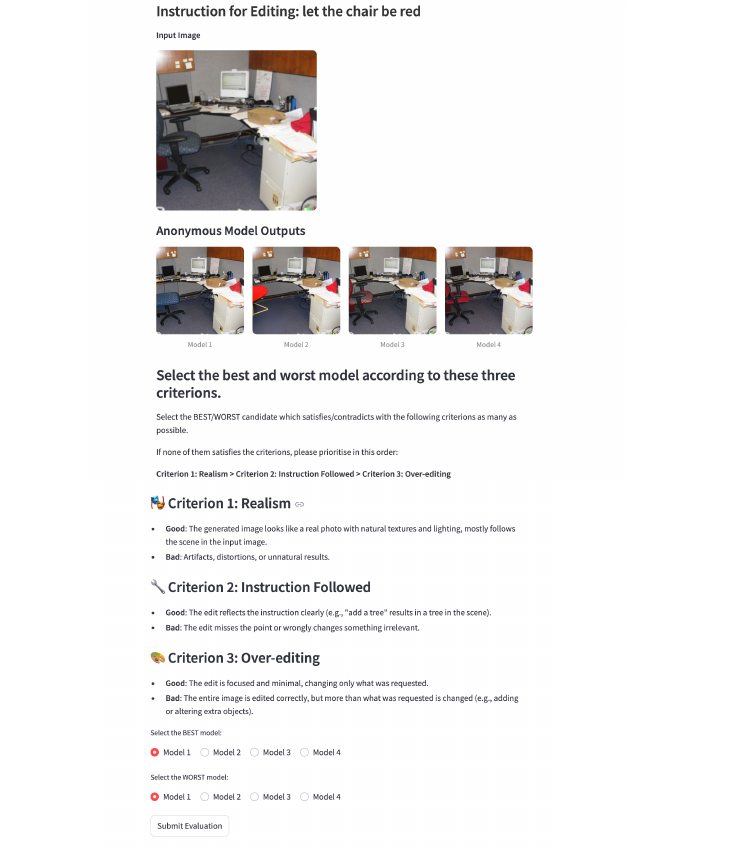}
    \caption{Instructions given to participants and the interface developed for conducting the evaluation.}
    \label{fig:human-eval-interface}
\end{figure*}

We conducted a human evaluation using a custom-built interface, with the full interface and instructions shown in Figure~\ref{fig:human-eval-interface}. A total of 14 participants were recruited, all of whom are PhD-level graduate students or higher. Participation was voluntary. Each participant was asked to evaluate 25 samples, which typically required 15–20 minutes to complete.

The evaluation process, including recruitment, instructions, and data processing and storage, followed our institution’s ethical guidelines for human subject research. All participants were informed of the purpose of the study and provided consent. No personally identifiable information was collected, and all data were stored and analysed in accordance with privacy standards.

\section{Safeguards}
\label{appendix:safeguards}

C-FDM performs observation prediction through image generation and, while its outputs are task-specific, we acknowledge that any generative model may carry potential for misuse. To mitigate these risks, we commit to the following safeguards upon release:

The model will be released solely for research purposes under a license that prohibits commercial use or any other harmful applications. The GitHub repository will include clear usage guidelines and terms of use, aligned with responsible AI principles.

We will include a disclaimer that the model is intended only for academic research in controlled environments. The datasets used for training are publicly available, action-centric image editing benchmarks that do not include sensitive or personally identifiable content.

Given the targeted nature of our model and the safeguards in place, we believe the risk of misuse is limited. Nonetheless, we encourage responsible use and welcome feedback from the community regarding potential improvements to safety.

\section{LLMs Usage Declaration}

We declare that the large language model (LLM) was only used to assist in minor tasks, including revising the manuscript for grammatical correctness, improving phrasing, and performing small technical implementations such as debugging code snippets. All scientific ideas, results, analyses, and conclusions presented in this paper are entirely the work of the authors.

\end{document}